%% file: evolq.tex
\PassOptionsToPackage{table}{xcolor}

\documentclass[10pt,twocolumn,letterpaper]{article}
\usepackage[accsupp]{axessibility} 
\usepackage{iccv}

\usepackage{subcaption}
\usepackage{graphicx}
\usepackage{amsmath}
\usepackage{amssymb}
\usepackage{booktabs}
\usepackage{algorithm}
\usepackage[noend]{algpseudocode}
\algnewcommand{\LineComment}[1]{\State \(\triangleright\) {\color{BlueViolet} #1 } }
\def\BState{\State\hskip-\ALG@thistlm}
\makeatother

\usepackage{pifont}

\DeclareMathOperator*{\argmin}{arg\,min} 

\usepackage[dvipsnames]{xcolor}
\usepackage{tcolorbox}

\definecolor{airforceblue}{rgb}{0.36, 0.54, 0.66}
\definecolor{darkblue}{rgb}{0.0, 0.0, 0.55}

\usepackage[pagebackref=true,breaklinks=true,letterpaper=true,colorlinks,bookmarks=false]{hyperref}

\usepackage[capitalize]{cleveref}
\crefname{section}{Sec.}{Secs.}
\Crefname{section}{Section}{Sections}
\Crefname{table}{Table}{Tables}
\crefname{table}{Tab.}{Tabs.}

\iccvfinalcopy 

\def\iccvPaperID{10131} 

\ificcvfinal\fi

\begin{document}

\title{Jumping through Local Minima: Quantization in the Loss Landscape of Vision Transformers}

\input{authors}

\maketitle
\ificcvfinal\thispagestyle{empty}\fi

\begin{abstract}
Quantization scale and bit-width are the most important parameters when considering how to quantize a neural network. Prior work focuses on optimizing quantization scales in a global manner through gradient methods (gradient descent \& Hessian analysis). Yet, when applying perturbations to quantization scales, we observe a very jagged, highly non-smooth test loss landscape. In fact, small perturbations in quantization scale can greatly affect accuracy, yielding a $0.5-0.8\%$ accuracy boost in 4-bit quantized vision transformers (ViTs). In this regime, gradient methods break down, since they cannot reliably reach local minima.
In our work, dubbed Evol-Q, we use evolutionary search to effectively traverse the non-smooth landscape.  
Additionally, we propose using an infoNCE loss, which not only helps combat overfitting on the small calibration dataset ($1,000$ images) but also makes traversing such a highly non-smooth surface easier. Evol-Q improves the top-1 accuracy of a fully quantized ViT-Base by  $10.30\%$, $0.78\%$, and $0.15\%$ for $3$-bit, $4$-bit, and $8$-bit weight quantization levels. Extensive experiments on a variety of CNN and ViT architectures further demonstrate its robustness in extreme quantization scenarios. Our code is available at \url{https://github.com/enyac-group/evol-q} .
\end{abstract}

\section{Introduction}
\label{sec:intro}

Quantization is a widespread technique for efficient neural network inference: reducing data precision from $32$ bits to $\leq$ $8$ bits is an effective approach to aggressively reduce model footprint and speed up computation. 
Network quantization is an extremely important tool for deploying models in cloud~\cite{vanholder2016efficient}  and edge settings~\cite{siddegowda2022neural}, where we aim to maximize accuracy while reducing the computational burden. We consider the post-training quantization (PTQ) setting, where there is access to a small ($\sim$$1,000$ image) calibration dataset but no access to the original training dataset. PTQ is an integral component of model deployment, when a carefully curated full-precision model is too expensive to retrain. Our work, Evol-Q, is a PTQ method for vision transformers which leverages four key observations:

\begin{figure}[t]
  \centering
  \begin{subfigure}{0.45\linewidth}
    \includegraphics[width=1\linewidth]
               {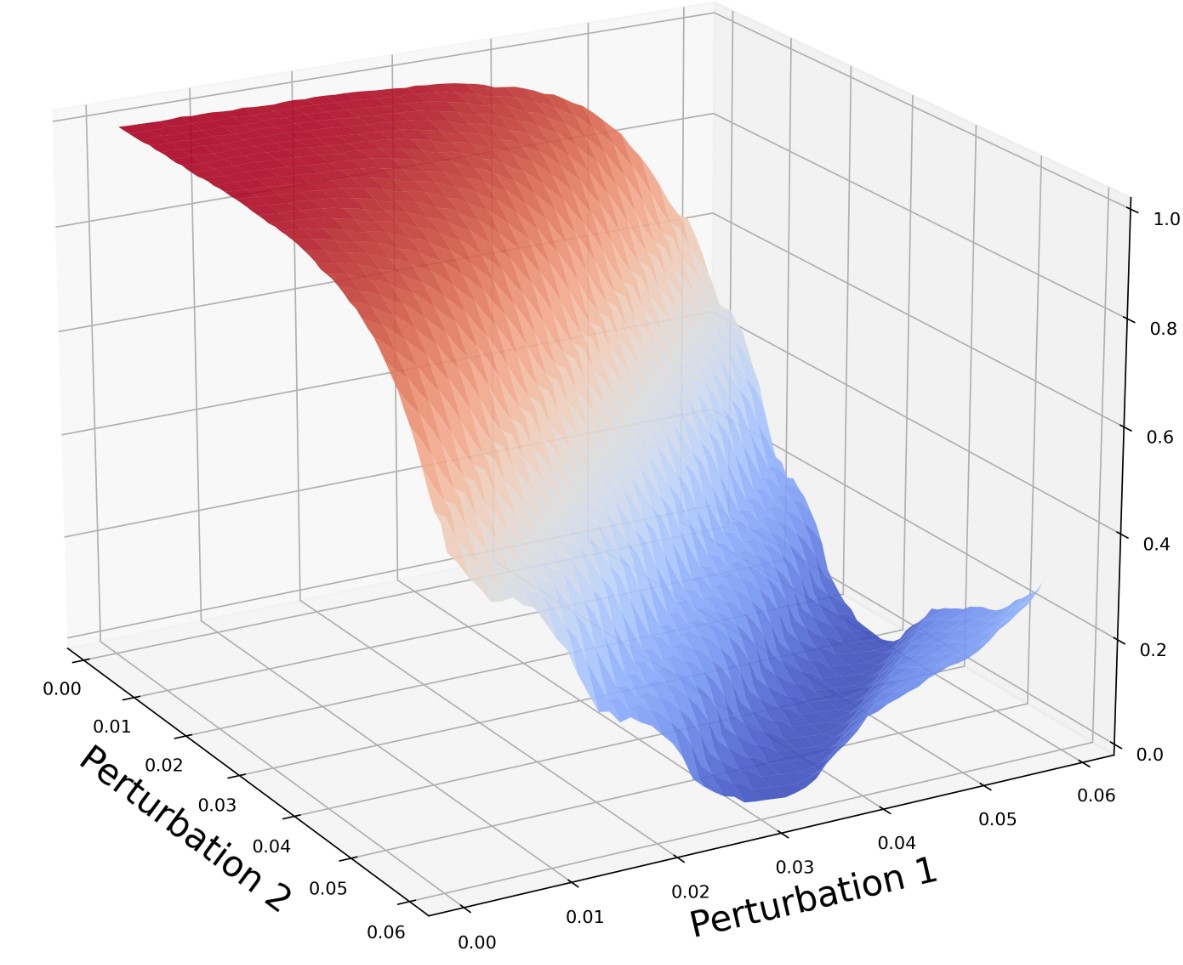}
    \caption{ResNet-18 Test Loss}
    \label{fig:cnn_loss}
  \end{subfigure}
  \hfill
   \begin{subfigure}{0.5\linewidth}
    \includegraphics[width=1\linewidth]
               {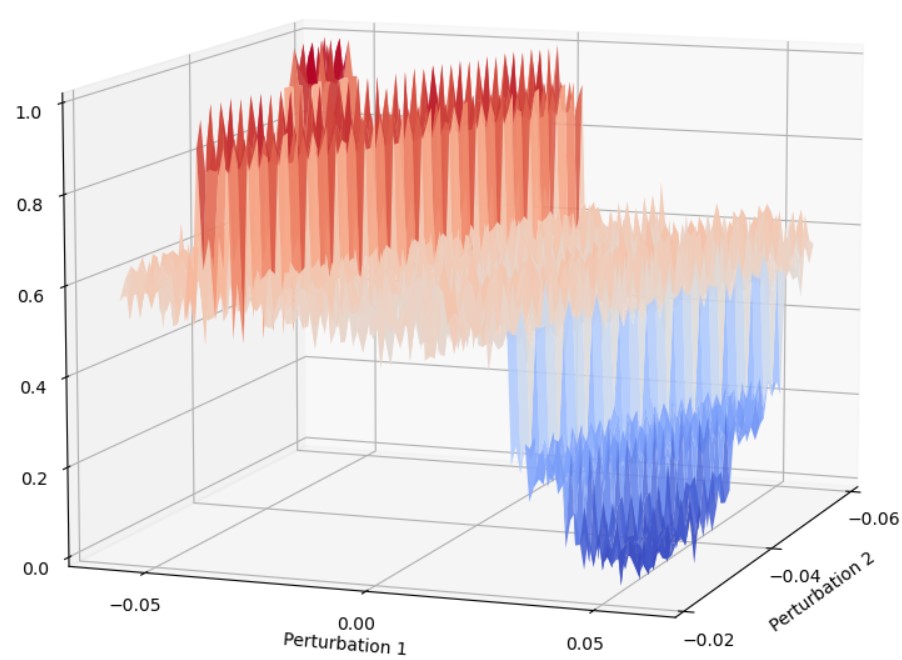}
    \caption{DeiT-Tiny Test Loss}
    \label{fig:vit_loss}
  \end{subfigure}
  \caption{We perturb along two basis vectors of one layer/block's quantization scales. The test loss landscape during perturbation is smooth in the CNN case (a), and highly non-smooth in the ViT case (b).}
  \label{fig:attn_grabber}
  
\end{figure}

\begin{enumerate}
    \item \textbf{Small perturbations in quantization scale can lead to significant improvement in quantization accuracy.} For example, small adjustments in a self-attention block's scale can induce a $\pm$ 1\% change in accuracy.
    \item As shown in \cref{fig:attn_grabber}, \textbf{quantized vision transformers (ViTs) have an extremely non-smooth loss landscape, particularly with respect to the perturbation in quantization scales}, making stochastic gradient descent a poor choice for optimization. We use evolutionary search to favor nearby local minima to significantly improve accuracy ($\sim$ $0.5-1\%$).
    \item In comparison to non-contrastive loss functions such as mean squared error, cosine similarity, and the KL divergence, \textbf{contrastive losses tend to smooth the loss landscape, as observed in our experiments and supported by recent work~\cite{fradkin2022robustness}}. This finding inspires the use of contrastive loss to further facilitate the quantization scale search process. Contrastive losses, specifically the infoNCE loss in this work, also helps in combating overfitting on the small calibration dataset by incorporating negative examples into the loss.
    \item \textbf{The Evol-Q framework generalizes to CNN quantization as well}, since the infoNCE loss provides a smoother landscape than other losses.
\end{enumerate}
 
Combining these observations, we devise a new optimization scheme to adjust the quantization scales of low bit-width ViTs.
Instead of using gradient descent to optimize all network parameters or estimating a noisy Hessian, we propose a series of cheap evolutionary search procedures to successively minimize the quantization error. Evol-Q injects small perturbations into the quantization scale at one layer and uses a global infoNCE loss to evaluate them. In the next section, we will show how prior work does not properly address the non-smooth ViT loss landscape.

\begin{figure*}
  \centering
  \begin{subfigure}{0.65\linewidth}
    \includegraphics[width=1\linewidth]
               {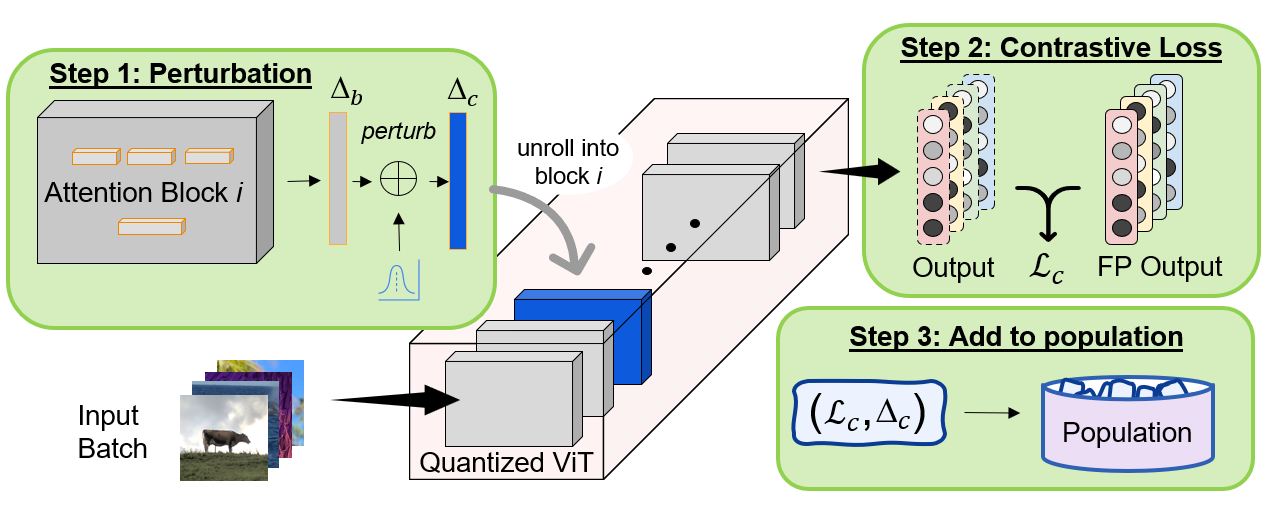}
    \caption{A single cycle of block-wise evolutionary search.}
    \label{fig:one-cycle}
  \end{subfigure}
  \hfill
   \begin{subfigure}{0.3\linewidth}
    \includegraphics[width=1\linewidth]
               {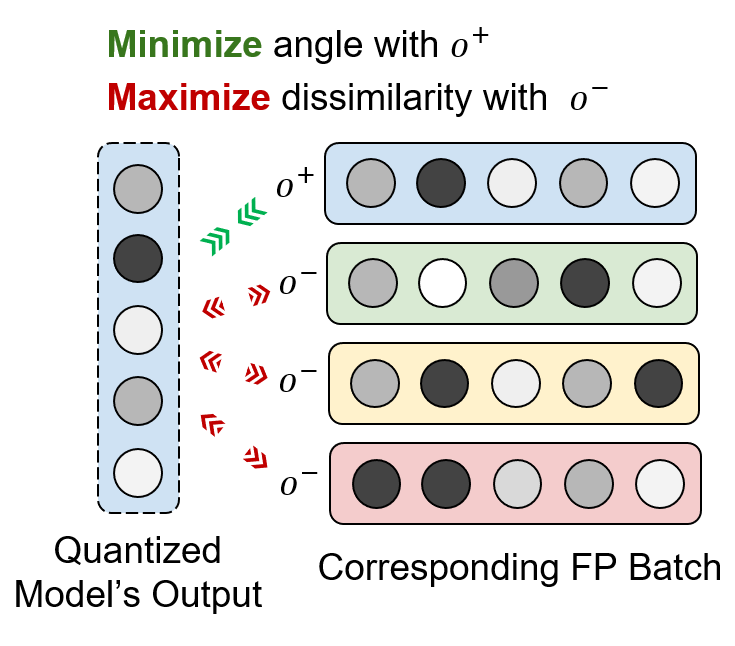}
    \caption{Visualization of the infoNCE Loss}
    \label{fig:loss-visualization}
  \end{subfigure}
  \caption{An overview of Evol-Q. On the left, we show one cycle completed on a single block. Each block has $C$ cycles of evolutionary search, and we perform $P$ passes over all blocks. On the right, we provide intuition for the infoNCE loss (Step 2), where we encourage similarity between the quantized and corresponding predictions while simultaneously maximizing dissimilarity between unlike predictions.}
  \label{fig:main-figure}
\end{figure*}

\section{Related Work}
\textbf{CNN Quantization:} When quantizing ViTs, it is natural to borrow techniques from CNN quantization and apply them to vision transformers. In the case of general quantization, one can consider either naive methods (such as MinMax~\cite{jacob2018quantization} quantization) or complex methods (such as gradient descent) to find the quantization scale.
Naive techniques include MinMax~\cite{jacob2018quantization}, Log2~\cite{cai2018deep}, or Percentile quantization, where we apply some statistical analysis on the values of a model tensor. While simple to implement, these methods can result in unacceptable accuracy degradation, particularly in the $3$-bit and $4$-bit case. This leads us to employ more advanced strategies to recoup some accuracy lost by quantization.

Complex methods involve techniques such as gradient descent\cite{nahshan2021loss}, Hessian analysis~\cite{li2021brecq}, or knowledge distillation~\cite{choi2020data} to maximize the quantized model's accuracy. In particular, many methods employ a layer-wise loss~\cite{wu2020easyquant, hubara2021accurate,   banner2019post}. A layer-wise loss can serve as a good proxy for a smooth global loss~\cite{nagel2020up}, yet a local loss is unlikely to resemble the ViT's highly non-smooth global landscape~\cite{bai2020binarybert}. While Hessian-based methods~\cite{li2021brecq} can utilize second order information, the ViT loss landscape resembles an ``egg carton" with a high density of extremal points. This space cannot be traversed well with only first and second-order gradient information. Moreover, BRECQ~\cite{li2021brecq} assumes the Hessian is positive semi-definite (PSD) which is a poor assumption for any non-smooth landscape.

\textbf{ViT Quantization:} Some work has already achieved very good accuracy on vision transformers. PTQ-for-ViT~\cite{liu2021post} learns a quantization scale for each layer by using two loss functions: (1) a ranking loss for each multi-head attention module; and (2) cosine similarity for the MLP layer. The layer-wise loss may achieve good accuracy, but with the non-smooth ViT landscape, this approach may end up at a local maximum and may be very sensitive to initialization. 
PTQ4ViT~\cite{yuan2022ptq4vit} also employs a Hessian-guided metric for guided global optimization similar to BRECQ~\cite{li2021brecq}. As we mentioned before, the PSD assumption on the Hessian breaks down for our ``egg carton''-shaped loss landscape. 
In contrast, PSAQ-ViT-V2~\cite{li2022psaq} uses a student-teacher MinMax game to minimize the KL divergence between the full precision and quantized models. The KL divergence is applied to kernel density estimations which are much more robust than Hessian-based or gradient-based techniques. We believe this technique can traverse the non-smooth ViT landscape, and despite being data-free, it has the best accuracy of all other methods we compare against.

We apply Evol-Q, our quantization scale learning method, on top of the FQ-ViT~\cite{lin2022fq} which incorporates Log2 quantization and an integer Softmax function for end-to-end $8$-bit quantization. FQ-ViT~\cite{lin2022fq} is surprisingly effective on ViTs, likely because Log2 quantization can compensate for the asymmetric distributions following the Softmax and GeLU layers.

Quantization-Aware Training (QAT) has shown impressive results on vision transformers~\cite{xu2022tervit, li2022q, li2022auto, li2022vit}, yet these methods consider training with the entire dataset rather than an unlabeled calibration dataset.

\textbf{Contrastive Loss:} In this work, we use a global infoNCE loss as our preferred loss function for both CNN and ViT quantization. Since we have such a small calibration dataset, the infoNCE loss helps generalize to the unseen test set. We are the first work to use a infoNCE loss in this manner. Prior work has combined quantization and self-supervised learning in a joint training framework~\cite{cao2022synergistic, fu2022contrastive} allowing for regularization from both the contrastive loss and a quantization loss in training. In particular, SSQL~\cite{cao2022synergistic} uses a joint SimSiam-$L_2$ loss during training to improve quantization for all bit-widths, whereas our method considers how 
the infoNCE loss, in conjunction with an evolutionary search, is used to traverse the highly non-smooth loss landscape and optimize the quantization scale for \textit{a specific bit-width at test time}. Moreover, SSQL~\cite{cao2022synergistic} uses the loss as regularization and not as a proxy for the test loss . Another work~\cite{shang2022network} applies contrastive learning to binarized ResNet and VGG models. They apply a layer-wise infoNCE loss, showing that it achieves good results for the 2-bit loss landscape of small CNNs. While a layer-wise loss is sub-optimal, the ability for the infoNCE loss to perform well on a binary loss landscape is great motivation for our work.

In summary, prior work has migrated CNN quantization techniques to ViT quantization without addressing the non-smooth ViT loss landscape. In the coming section, we present Evol-Q as an effective solution to this problem.
\section{ The Evol-Q Framework}
In the non-smooth ViT quantization landscape, first and second-order gradient methods are not effective and cannot handle the large number of local minima. In such a regime, small perturbations in quantization scale can lead to a significant boost in accuracy. We apply evolutionary search using an infoNCE loss to evaluate these perturbations, enabling Evol-Q to traverse a non-smooth surface and minimize overfitting on the calibration dataset.

We begin with an overview of uniform quantization and then dive into the core components of our method: traversing perturbations using evolutionary search and evaluating them using an infoNCE loss.

\subsection{Uniform, End-to-End Quantization}
We consider uniform quantization, where full precision values are mapped to a uniform range based on the quantization bit-width (b). Uniform quantization is formally defined as:

\begin{equation}
  Q(\textbf{x}, \delta, b) = \texttt{clip}(\Big{\lfloor} \frac{\textbf{x}}{\delta} \Big{\rceil}, -2^{b-1}+1, 2^{b-1}-1)
  \label{eq:quant}
\end{equation}
where $\textbf{x}$ is a full-precision vector and  $\delta$ is the quantization scale. We can generalize this idea to a tensor \textbf{X} and a corresponding quantization vector $\Delta$  (\textit{e.g,} each element in $\Delta$ corresponds to a channel in channel-wise quantization). In our framework, we learn these initial quantization parameters using a fast, layer-wise framework such as FQ-ViT~\cite{lin2022fq}, and then use evolutionary search to adjust the scales of each attention block's projection layers.

\subsection{Where to Perturb?}
A ViT model~\cite{dosovitskiy2020image} consists of $12$ multi-head self-attention (MHSA) blocks stacked on top of each other. Each block  applies the following transformation on a given query (Q), key (K), and value (V):
\begin{equation}
    MHSA(Q,K,V) = \texttt{concat}(H_0, H_1, \dots H_N) W^O
  \label{eq:mhsa}
\end{equation}
where each attention head $H_i$ is:
\begin{equation}
    H_i(Q,K,V) = \texttt{softmax}(\frac{(W^QQ) (W^K K)^{T}}{\sqrt{d_k}})\cdot W^V V
  \label{eq:hi}
\end{equation}

Our method applies end-to-end quantization meaning that for each attention block we quantize all $3N + 1$ weight tensors and $6N + 1$ intermediary activations where N is number of heads. The quantization scales of all weights and activations can be concatenated and viewed as the vector $\Delta$. This stacked vector is very important for understanding our algorithm -- we can perturb the scales for all weights and activations simultaneously by perturbing $\Delta$. We perturb by sampling from a uniform ball centered around $\Delta$:
\begin{equation}
  \boldsymbol\Delta_{\texttt{new}} \sim \mathcal{U}(\boldsymbol\Delta, -\epsilon, \epsilon)
  \label{eq:perturb}
\end{equation}
where $\epsilon$ controls the size of the uniform ball. Perturbations within a small $\epsilon$-ball ($\epsilon \approx 10^{-4}$) yields a change in accuracy of $\pm 1\%$ top-1 accuracy for $4$-bit quantization. This is the same order of magnitude used to compare a variety of quantization methods, further illustrating that small perturbations matter considerably for quantization performance.

We will show in sequel how evolutionary search can quickly evaluate multiple perturbations and choose which local minima to hop into.

\subsection{Global Search for Quantization Scales}
As previously mentioned, we perturb the quantization scales ($\Delta$) for one attention block at a time. If we perturb too many scales at once, we end up traversing a very high dimensional search space and the number of iterations for evolutionary search convergence increases exponentially. Of course, gradient descent is often effective for such large search spaces, but first-order methods will not work well in our non-smooth loss regime. After partitioning our search space in a block-wise manner, we find evolutionary search to perform well and achieve acceptable convergence times (on par with the runtime of other PTQ methods).

We apply a small evolutionary search algorithm for each attention block (shown in \cref{fig:main-figure}), and repeat this for all blocks in the model. A block's evolutionary search algorithm has $C$ cycles, where each cycle spawns a perturbation. The search population is initially set to $|K|$ copies of the original scale, and each cycle progressively updates the population of scales. During one cycle, we choose a parent scale from the population and a perturbation (child scale) is then spawned from the uniform distribution parameterized by the parent scale (see \cref{eq:perturb}). The child scales are then placed into the population for the next cycle. We evaluate each child scale using the fitness function defined in \cref{alg:fitness} which applies a global infoNCE loss. The block-wise search algorithm is shown in \cref{alg:joint-evol}. We refer the reader to evolutionary computing literature~\cite{eiben2015introduction} for more details on parent, child, and fitness function.

In summary, our block-wise evolutionary search consists of $P$ passes over all attention blocks. For each attention block $b$, we apply a small evolutionary algorithm for $C$ cycles, meaning that each block's quantization scale is adjusted $P \times C$ times. A full enumeration of the search settings are in \cref{tab:settings}. 

\begin{algorithm}[t]
\caption{Block-wise Evolutionary Search}\label{alg:joint-evol}
\hspace*{\algorithmicindent} \textbf{Input:} Calibration Dataset $D_{C}$ \\
\hspace*{\algorithmicindent} \hspace{3mm} Quantized Model $M_{Q}$, Full Precision Model $M_F$ \\
\hspace*{\algorithmicindent} \hspace{3mm} Number of passes $P$, cycles $C$ \\
\hspace*{\algorithmicindent} \hspace{3mm} Population size $K$, Sample size $S$
\begin{algorithmic}[1]
\For{ passes in 0 : $P$}
\LineComment{traverse all attention blocks}
\For{each attention block $b$}
    \LineComment{sub-problem illustrated in \cref{fig:main-figure}}
    \State pop $\gets$ [\hspace{1mm} ]
    \LineComment{init population with $K$ perturbations}
    \While {$|$ pop $|$ $<$ $K$ }
        \LineComment{\texttt{Fitness} defined in \cref{alg:fitness}}
        \State fitness $\gets$ \texttt{Fitness}($b$, $M_Q$, $M_F$, $D_{C}$)
        \State pop\texttt{.insert(($\boldsymbol\Delta_{\texttt{b}}$, fitness))}
    \EndWhile
    \LineComment{Iterate to find best child $\Delta$}
    \For{ cycles in 0 : $C$}
        \LineComment{From population, sample a parent $\Delta$}
        \State samples $\gets$ [\hspace{1mm} ]
        \While {$|$ samples $|$ $<$ $S$ }
            \State samples $\gets$ \texttt{RandomElement}(pop)
        \EndWhile
        \State parent $\gets$ \texttt{BestFitness}(samples)
        \State
        \LineComment{Generate child and add to population}
        \State $\boldsymbol\Delta_{\texttt{child}}$ $\gets$ \texttt{Perturb}($\boldsymbol\Delta_{\texttt{parent}}$)
        \State fitness $\gets$ \texttt{Fitness}($D_C$, $M_Q$, $M_F$ )
        \State pop\texttt{.insert}( ($\boldsymbol\Delta_{\texttt{child}}$, fitness ) )
        
        \LineComment{remove candidate with lowest fitness}
        \State pop.\texttt{DeleteDead}()
    \EndFor
    \LineComment{Choose best perturbation as b's final $\Delta$}
    \State $\boldsymbol\Delta_{\texttt{b}}$ $\gets$ \texttt{BestFitness}(pop)
\EndFor
\EndFor
\State \Return $M_{Q}$
\Comment model with updated scales
\end{algorithmic}
\end{algorithm}

\subsection{The infoNCE Loss for Scale Search}
\label{sec:fitness}
If we apply evolutionary search on the test loss landscape, we can quickly traverse the space of local minima and find the best quantization scale. Unfortunately, the test loss is not known to us, and the small calibration dataset may produce a loss landscape that is different from the true loss landscape. We find that the infoNCE loss can incorporate negative samples to reduce the model's tendency to overfit on positive bias.

The infoNCE loss is a common contrastive loss function used in self-supervised learning to smooth the loss landscape~\cite{fradkin2022robustness}, prevent representation collapse \cite{grill2020bootstrap} and encourage discrimination between the target representation and a set of negative examples. We find the infoNCE loss to be very effective to prevent overfitting of a \textit{quantized network's} representation in the same way that it is used to develop richer representations in a self-supervised setting~\cite{fradkin2022robustness} (see supplementary materials for details). Inspired by Chen et al.~\cite{chen2021empirical}, we use the infoNCE~\cite{oord2018representation} loss:
\begin{equation}
  \mathcal{L}_{c} = - \log \frac{\exp(p \cdot o^{+} / \tau) }{\exp(p \cdot o^{+} / \tau) + \sum_{o^{-}} \exp(p \cdot o^{-} / \tau)}
  \label{eq:infoNCE}
\end{equation}

where $p$ is the prediction of the \textbf{quantized model}. The infoNCE loss is \textbf{sampled within the full-precision model's batch}, where $o^{+}$ is the corresponding prediction to $p$, and $o^{-}$ is a prediction of other images in the same batch. In \cref{alg:fitness}, we show how the infoNCE is evaluated in the fitness function for Evol-Q.
\setlength{\textfloatsep}{0.1cm}
\setlength{\floatsep}{0.1cm}

\begin{algorithm}
\caption{Fitness Function}\label{alg:fitness}
\hspace*{\algorithmicindent} \textbf{Input:} calibration dataset $D_C$ \\
\hspace*{\algorithmicindent} \hspace{3mm} quantized model $M_Q$, full precision model $M_F$
\begin{algorithmic}[1]
\For{ batch in $D_{C}$}
\State p = $M_Q$(batch)
\State o = $M_F$(batch)
\State score += \texttt{infoNCE}(p, o)
\Comment{\cref{eq:infoNCE}}
\EndFor
\State \Return score / \texttt{size}($D_C$)
\end{algorithmic}
\end{algorithm}

\section{Results}
In the following section, we present results on a variety of vision transformers and show the consistency of our method under standard $8$-bit quantization and in extreme quantization schemes ($3$-bit and $4$-bit weights). We present results for end-to-end quantization, where all weights and activations are quantized.
\subsection{Setup}
In our post-training (PTQ) setup, the calibration dataset is $1,000$ randomly sampled images from the ImageNet training set. Experiments are conducted on ImageNet (ILSVRC2012), and we evaluate top-1 accuracy for a variety of ViT model families.
For Evol-Q, the initial quantized model ($M_Q$ in \cref{alg:joint-evol}) is generated using FQ-ViT, and our method perturbs its quantization scales to yield better performance. FQ-ViT is an end-to-end quantization framework that uses MinMax~\cite{jacob2018quantization} for weight quantization and Log2~\cite{cai2018deep} for activation quantization. We refer to FQ-ViT and our code for other quantization settings. The Evol-Q search parameters (from \cref{alg:joint-evol}) are in \cref{tab:settings}.

\begin{table}[h]
    \centering
    \begin{tabular}{ c c|c }
    \toprule
         passes & $P$ & 10 \\
         population size & $K$ & 15 \\
         cycles & $C$ & 3 \\
         samples & $S$ & 10 \\
         mutation range & $\epsilon$ & $10^{-3}/10^{-4}$ \\
    \bottomrule
    \end{tabular}
    \caption{Block-wise evolutionary search settings. The mutation range is $10^{-3}$ for 8W8A, and $10^{-4}$ for 4W8A and 3W8A.}
    \label{tab:settings}
\end{table}

\subsection{8-bit Quantization}
We compare our standard 8-bit quantization with state-of-the-art methods in \cref{tab:8w8a}.  Evol-Q improves over existing \textit{end-to-end} quantization techniques by $0.1\%$, $1.2\%$, and $0.15\%$ for DeiT-Small, DeiT-Base, and ViT-Base, respectively. 

\begin{table}[h]
  \centering
  \rowcolors{2}{gray!25}{white}
  \begin{tabular}{| c | c  c  c  c |}
    \hline
    \multicolumn{5}{|c|}{8-bit weights, 8-bit activations (8W8A)} \\
    \hline
    Method & DeiT-T & DeiT-S & DeiT-B & ViT-B \\
    \hline
    PSAQ-ViT & 71.56 & 76.92 & 79.10 & 37.36 \\
    PTQ4ViT  & - & 79.47 & 81.48 & 84.25 \\
    FQ-ViT  & 71.61 & 79.17 & 81.20 & 83.31 \\
    PSAQ-ViT-V2\textsuperscript{\textdagger} & \textbf{72.17} & 79.56 & 81.52 & -  \\
    Evol-Q (ours) & 71.63 & \textbf{79.57} & \textbf{82.67} & \textbf{84.40} \\
    \hline
  \end{tabular}
  
  \footnotesize{\textsuperscript{\textdagger} Does not quantize Softmax/GeLU layers}
  \caption{Top-1 Accuracy on ImageNet using 8W8A quantization. -T, -S, \& -B refer to Tiny, Small, and Base models respectively.}
  \label{tab:8w8a}
\end{table}

We also compare with PSAQ-ViT-V2~\cite{li2022psaq} and find that it outperforms our method by $0.5\%$ for DeiT-Tiny. However, PSAQ-ViT-V2 is not a full end-to-end quantization method since it \textit{does not quantize the activations} following the Softmax and GeLU layers. These activations are typically very sensitive to quantization and are often maintained at full precision. We applied Evol-Q on an end-to-end quantized model, so we are forced to quantize the post-Softmax/GeLU activations. We leave it to future work to apply our technique on top of PSAQ-ViT-V2, but expect similar improvements to what we achieved with FQ-ViT.

\subsection{4-bit Quantization}

Moving from $8$-bit to $4$-bit weight quantization, we see an accuracy degradation of about $2-5\%$ across all models. In \cref{tab:4w8a}, Evol-Q performs similarly to what is shown for 8-bit quantization. In particular, we still see improvement for DeiT-Small, DeiT-Base, and ViT-Base, but now the top-1 accuracy improvement is $0.13\%$, $0.16\%$, and $0.77\%$, respectively.

\begin{table}[h]
  \centering
  \rowcolors{2}{gray!25}{white}
  \begin{tabular}{| c | c  c  c  c |}
    \hline
    \multicolumn{5}{|c|}{4-bit weights, 8-bit activations (4W8A)} \\
    \hline
    Method &  DeiT-T & DeiT-S & DeiT-B & ViT-B \\
    \hline
    PSAQ-ViT & 65.57 & 73.23 & 77.05 & 25.34 \\
    PTQ4ViT  & - & - & 64.39 & - \\
    FQ-ViT  & 66.91 & 76.93 & 79.99 & 78.73 \\
    PSAQ-ViT-V2\textsuperscript{\textdagger}  & \textbf{68.61} & 76.36 & 79.49 & -  \\
    Evol-Q (ours) & 67.29 & \textbf{77.06} & \textbf{80.15} & \textbf{79.50} \\
    \hline
  \end{tabular}
  \footnotesize{\textsuperscript{\textdagger} Does not quantize Softmax/GELU layers}
  \caption{Top-1 Accuracy on ImageNet using 4W8A quantization.}
  \label{tab:4w8a}
\end{table}
\subsection{3-bit Quantization}

We report 3-bit quantization results in \cref{tab:3w8a} to show that Evol-Q extends to more extreme quantization scenarios. In particular, Evol-Q improves accuracy over FQ-ViT by $10.3\%$ for ViT-Base and $3.7\%$ for DeiT-Tiny.

\begin{table}[h]
  \centering
  \rowcolors{2}{gray!25}{white}
  \begin{tabular}{| c |  c  c  c  c |}
    \hline
    \multicolumn{5}{|c|}{3-bit weights, 8-bit activations (3W8A)} \\
    \hline
    Method &  DeiT-T & DeiT-S & DeiT-B & ViT-B \\
    \hline
    FQ-ViT &  35.79 & 60.58 & 72.11 & 55.33 \\
    Evol-Q (ours)  & \textbf{39.45} & \textbf{61.16} & \textbf{72.41} & \textbf{65.63} \\
    \hline
  \end{tabular}
  \caption{Top-1 Accuracy on ImageNet with 3W8A quantization.}
  \label{tab:3w8a}
\end{table}
By reducing the precision from $32$ to $3$ bits, we achieve a $>$10X reduction in memory footprint while still maintaining a reasonable accuracy for DeiT-Base. We refer to supplementary materials for ablations on using OMSE~\cite{choukroun2019low} and bias correction~\cite{banner2019post} for 3W8A which dramatically improves our method's performance.

\subsection{Extending to Swin \& LeViT Models}
We run our experiments on additional model families to ensure that our method is applicable to different types of attention blocks.
Swin transformers~\cite{liu2021swin} have the same macro-architecture as DeiT and ViT models, with the exception that the Swin transformer block is a windowed attention. We see results for 4-bit Swin transformers in \cref{tab:swin}.

\begin{table}[h]
  \centering
  \rowcolors{2}{gray!25}{white}
  \begin{tabular}{| c |  c  c  c |}
    \hline
    Method &  Swin-T & Swin-S & Swin-B \\
    \hline
    PSAQ-ViT & 71.79 & 75.14 & - \\
    PSAQ-ViT-V2\textsuperscript{\textdagger} & 76.28 & 78.86 & - \\
    FQ-ViT & \textbf{80.73} & 82.13 & 82.73 \\
    Evol-Q (ours) & 80.43 & \textbf{82.63} & \textbf{83.07} \\
    \hline
  \end{tabular}
  
  \footnotesize{\textsuperscript{\textdagger} Does not quantize Softmax/GELU layers} \\
  \caption{Top-1 Accuracy for 4W8A Swin Models.}
  \label{tab:swin}
\end{table}
Prior quantization techniques do not consider LeViT models, a family of ViTs that improve the inference speed of existing transformers. Using a longer convolutional backbone, they can achieve similar results to classic ViTs while also reducing the complexity of the transformer blocks in favor of ResNet-like stages. We include the LeViT family in our experiments to illustrate how our method can be extended beyond the standard block size. We can see 4W8A results for the LeViT model family in \cref{tab:levit}. Across the board, we see a significant improvement in LeViT quantization as compared to FQ-ViT, our baseline. 

\begin{table}[h]
    \centering
    \rowcolors{2}{gray!25}{white}
    \begin{tabular}{|c|c c c c |}
        \hline
        & LeViT & LeViT & LeVit & LeViT \\
        Method & -128S & -192 & -256 & -384 \\
        \hline
        FQ-ViT & 14.90 & 17.00 & 61.33 & 64.60 \\
        Evol-Q (ours) & \textbf{29.20} & \textbf{30.37} & \textbf{64.57} & \textbf{69.50} \\
        \hline
    \end{tabular}
    \caption{Top-1 Accuracy for 4W8A LeViT models. }
    \label{tab:levit}
\end{table}
In fairness, we have not applied any other techniques to boost LeViT's accuracy (doing so may inflate our method's improvement), so we leave it to future work to incorporate other quantization techniques on top of our framework.
\section{Analysis}
In the following section, we support the three observations set forth in the introduction. First, we show how \textbf{the non-smooth ViT loss landscape compares with the CNN one}, and discuss how a variety of loss functions perform in the Evol-Q framework. Next, we visualize the layer-wise weight distributions to illustrate how small perturbations can yield a significant jump in accuracy. Finally, we report runtime and various ablations to contextualize our method in the broader space of quantization techniques. For more ablations and analysis, please refer to the supplementary materials.
\begin{figure}[h]
  \centering
  \begin{subfigure}{0.45\textwidth}
    \includegraphics[width=0.95\linewidth]
               {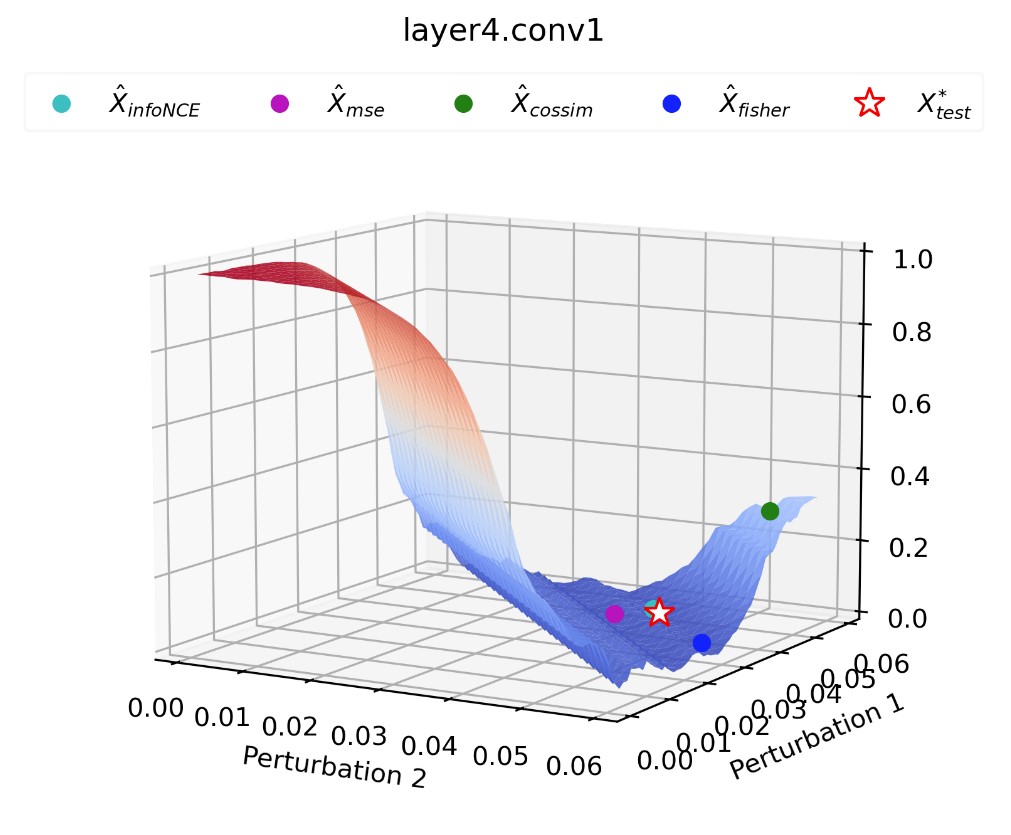}
    \caption{Resnet-18 loss landscape}
    \label{fig:cnn-loss-with-labels}
  \end{subfigure}
  \hfill
   \begin{subfigure}{0.45\textwidth}
    \includegraphics[width=0.95\linewidth]
               {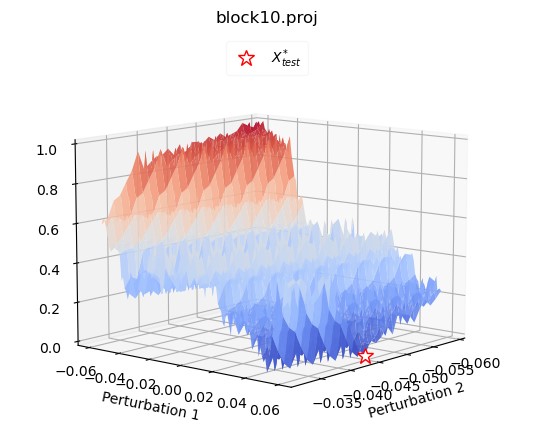}
    \caption{DeiT-Tiny loss landscape}
    \label{fig:vit-loss-with-labels}
  \end{subfigure}
  \caption{A comparison of the test loss landscapes for 4-bit quantized CNNs and ViTs. In \cref{fig:cnn-loss-with-labels}, we show how small perturbations in the $4^{th}$ convolutional layer yields a smooth test loss landscape. In \cref{fig:vit-loss-with-labels}, we apply perturbations to attention block \#10 and the resultant loss landscape is highly non-smooth.}
  \label{fig:test-losses}
\end{figure}
\subsection{The Test Loss Landscape of ViTs}
In \cref{fig:test-losses}, we show that perturbing quantization scale yields a smooth test loss landscape for ResNet-18 and a jagged, non-smooth landscape for DeiT-Tiny. This loss landscape illustrates how the test loss is related to the $\Delta_{10}$ scale at block \#10. \textbf{We perturb along two basis vectors for the $\Delta_{10}$ and observe the effect on the test loss.} The DeiT-Tiny loss landscape is very complex and highly non-linear, whereas the ResNet-18 landscape is comparatively smooth. Intuitively, we hypothesize that the presence of many GeLUs and Softmax functions induces in the DeiT loss landscape many more extreme points than in the ResNet loss landscape. 

For the smooth landscape in \cref{fig:cnn-loss-with-labels}, the infoNCE loss is clearly optimal since it is closest to the global minimum. If we plot the MSE, Cosine (Cossim), and Fisher loss landscapes, we find that they are not as smooth as the infoNCE loss in the CNN case (see supplementary materials, Sec A). The infoNCE loss helps to provide a smoother loss with respect to the calibration dataset by incorporating negative samples to reduce bias~\cite{fradkin2022robustness}.

In the non-smooth landscape, \cref{fig:vit-loss-with-labels}, the global minimum is very hard to find. In fact, proximity to the global loss in this landscape is not a good indicator of loss function quality, since there are many local maxima in close proximity (with an $\epsilon$- ball) of the global minimum. In \cref{sec:loss-func}, we provide an empirical justification that the infoNCE loss prevents overfitting and is superior to other loss functions.

\begin{figure}
    \centering
    \includegraphics[width=1\linewidth]
           {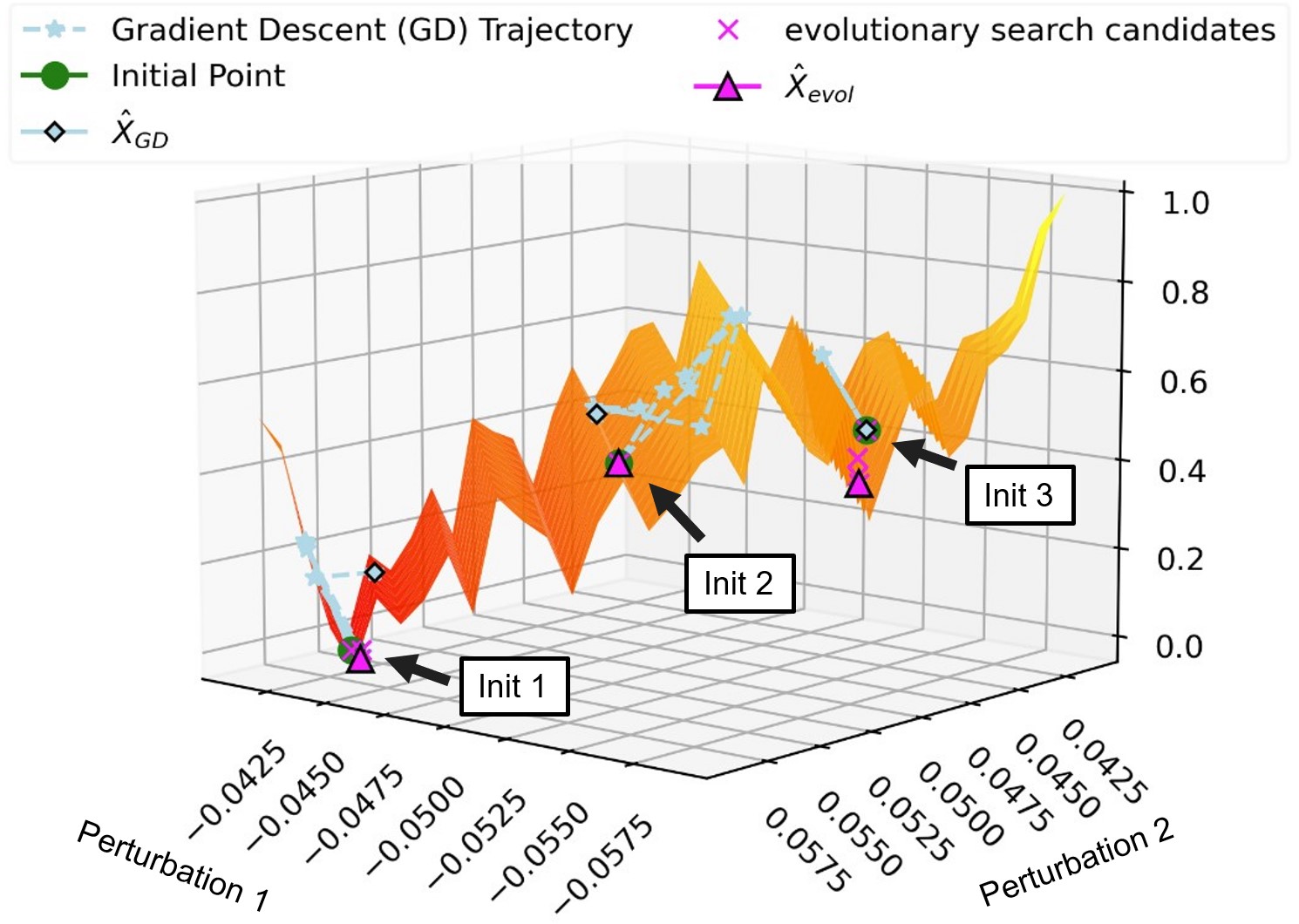}
    \caption{A zoomed in section of the landscape in \cref{fig:vit-loss-with-labels}, where we perform gradient descent and evolutionary search for three initial points. We show the solutions of evolutionary search ($\hat{X}_{evol}$ ) and gradient descent ($\hat{X}_{GD}$) after 10 iterations. }
    \label{fig:evol_v_grad}
\end{figure}

\begin{figure*}[t]
\centering
\begin{subfigure}{0.18\linewidth}
    \includegraphics[width=1\linewidth]{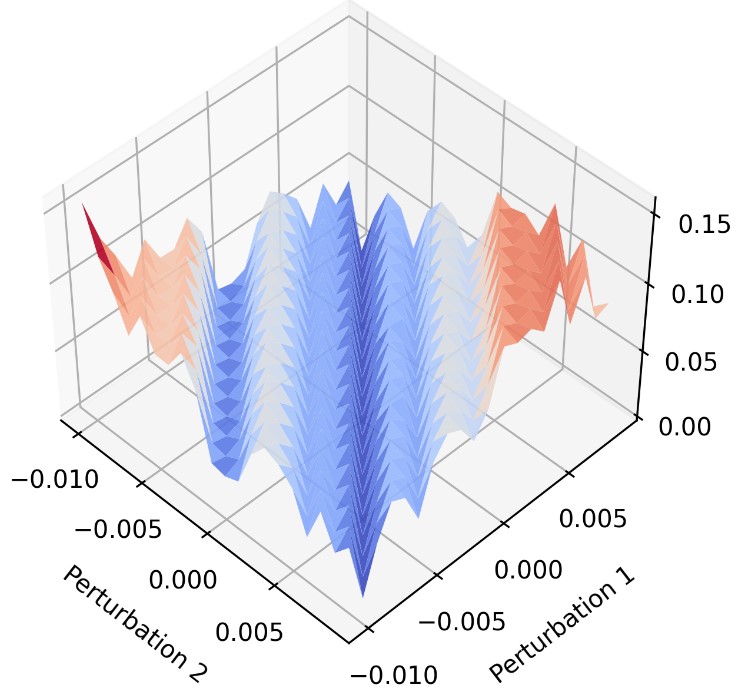}
    \caption{Query, Key, Value}
    \label{fig:qkv_min}
\end{subfigure}
\hfill
\begin{subfigure}{0.18\linewidth}
    \includegraphics[width=1\linewidth]{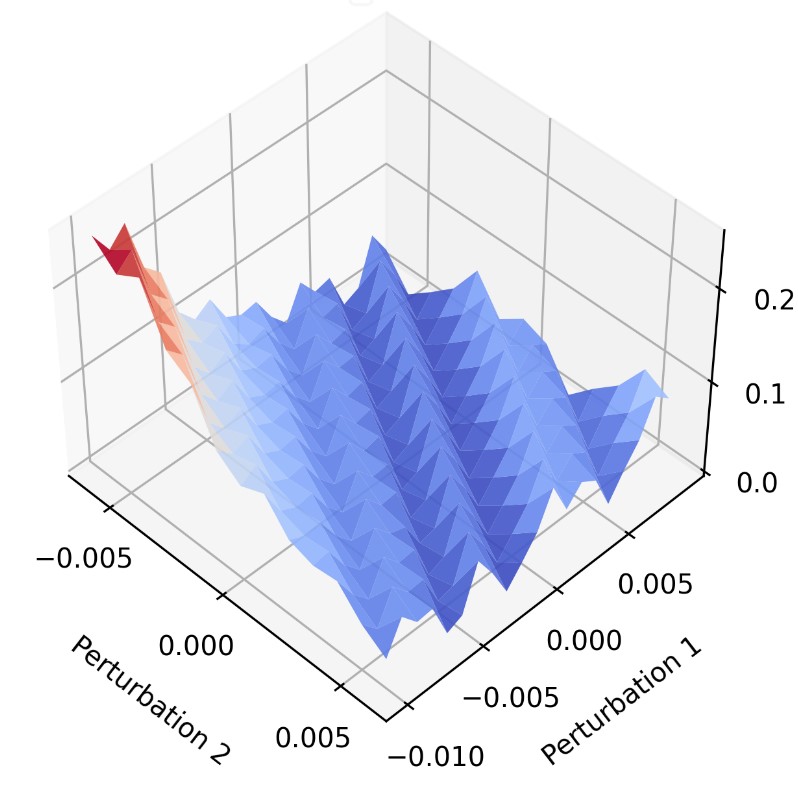}
    \caption{Projection Layer}
    \label{fig:proj_min}
\end{subfigure}
\hfill
\begin{subfigure}{0.18\linewidth}
    \includegraphics[width=1\linewidth]{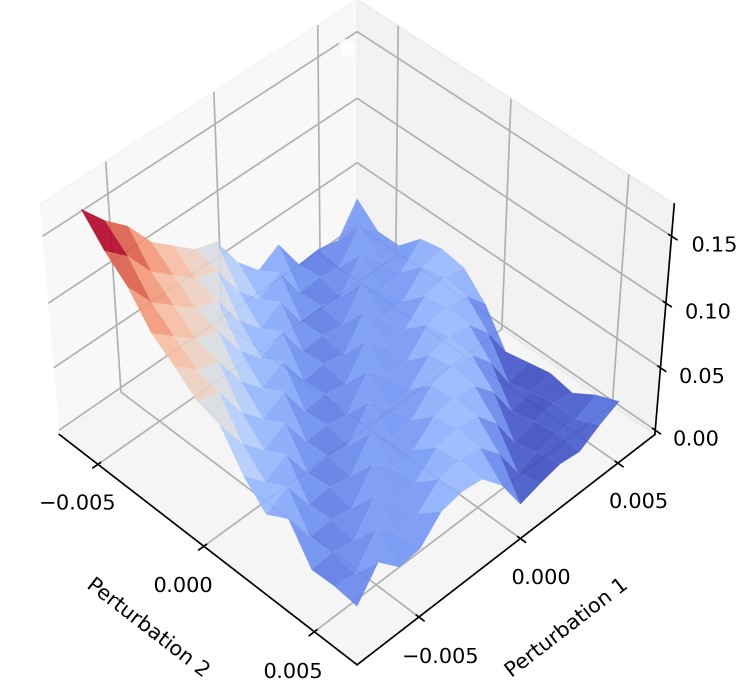}
    \caption{FC Layer \#1}
    \label{fig:fc1_min}

\end{subfigure}
\hfill
\begin{subfigure}{0.18\linewidth}
    \includegraphics[width=1\linewidth]{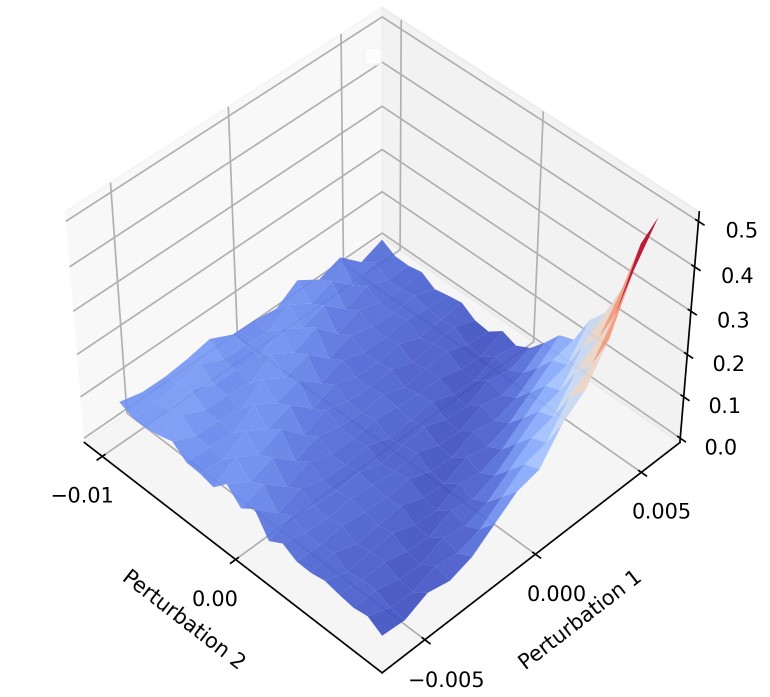}
    \caption{FC Layer \#2}
    \label{fig:fc2_min}

\end{subfigure}

\caption{Loss Landscapes for the 4-bit quantized QKV, Projection, and Fully Connected (FC) layers in self-attention block \#5. We perturb the the quantization scale along two basis vectors (Perturbation 1 \& 2) to visualize the loss landscape. These landscapes capture a zoomed in region around the global minimum of the full landscape. The FC layers exhibit relative smoothness around the global minimum whereas the QKV \& Projection layers are not easily traversible. The Projection layer is particularly difficult for gradient methods because it has 4 deep minima in close proximity to the global minimum. }
\label{fig:loss_land}
\end{figure*}
\subsection{Gradient Descent vs. Evolutionary Search}
In \cref{fig:evol_v_grad}, we show how evolutionary search finds candidates close the local minima, whereas gradient descent breaks down the ViT loss landscape. We show three initial points, where gradient descent either oscillates (init 3) or does not converge to a local minima (init 1 and  2). In contrast, evolutionary search generates a candidate perturbation and steps in the direction of the best candidate. We find that evolutionary search is very good at finding the closest local minima, which is sufficient to get an accuracy boost of $\sim 0.4\%$ in this loss landscape.

In Table \ref{tab:optimizers}, we show quantitatively that gradient-based optimizers underperform in comparison to evolutionary search for the same block-wise setting as in Evol-Q. We believe that non-smoothness at the block level is what makes these gradient-based techniques ineffective.
\begin{table} [h]
  \centering
  \resizebox{0.86\columnwidth}{!}{%
  \begin{tabular}{| c |  c  c  c  c |}
    \hline
    Method &  DeiT-T & DeiT-S & DeiT-B & ViT-B \\
    \hline
    SGD & 71.57 & 79.25 & 81.24 & 83.40 \\
    Adam & 71.29 & 79.25 & 81.24 & 83.25\\
    AdamW & 71.37 & 79.00 & 81.30 & 83.36 \\
    Evol-Q (ours)  & \textbf{71.63} & \textbf{79.57} & \textbf{82.67} & \textbf{84.40} \\
    \hline
  \end{tabular}
  }
  \vspace{-2mm}
  \caption{Comparison with gradient-based optimizers using 8-bit weights, 8-bit activations (8W8A).}
  \label{tab:optimizers}
  \vspace{-1mm}
\end{table}

\subsection{Loss Function Choice}
\label{sec:loss-func}
\begin{figure}
  \centering
\includegraphics[width=1\linewidth]
               {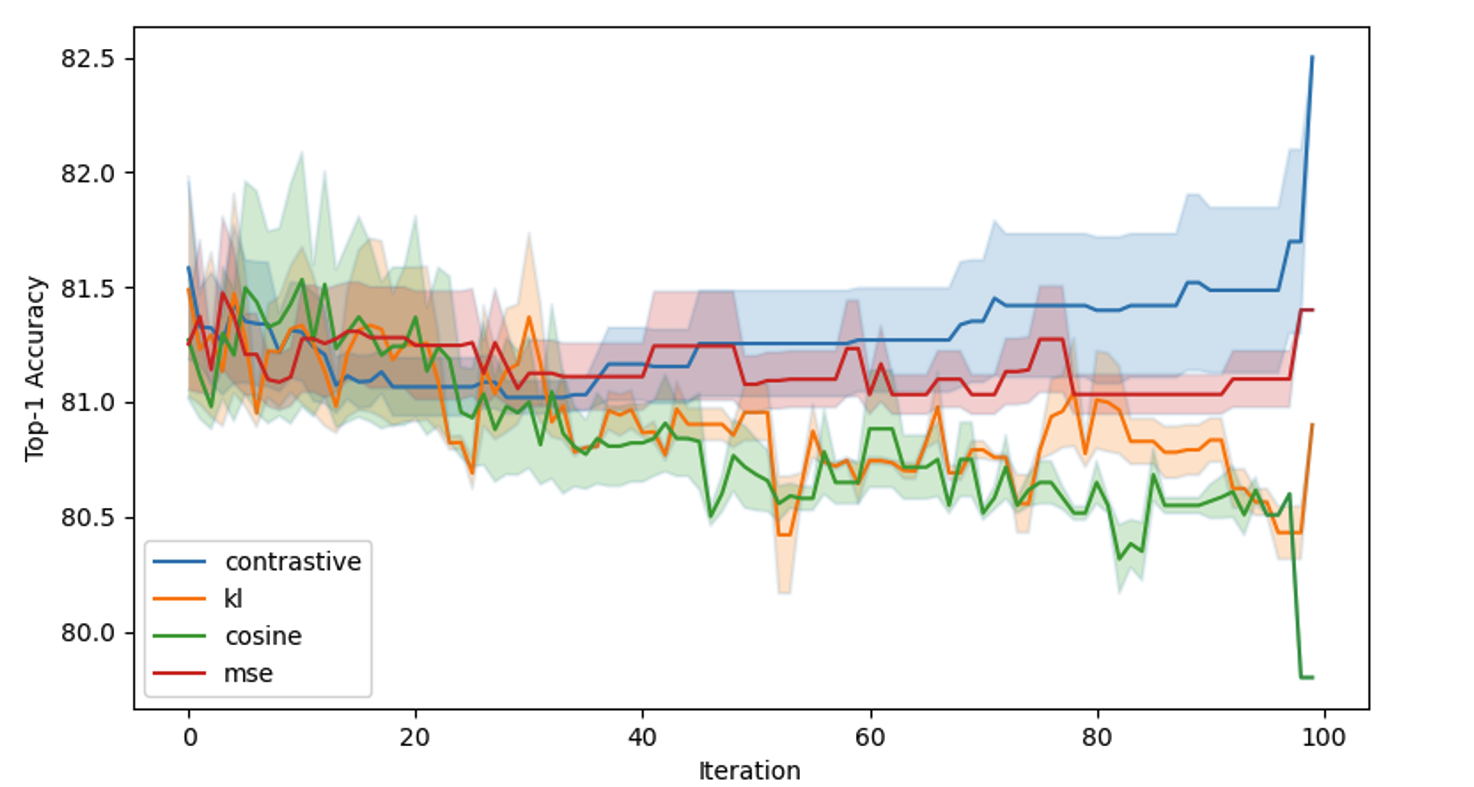}
\caption{Comparing four loss functions in the Evol-Q framework on ViT-Base. The infoNCE  (contrastive) loss prevents overfitting to the calibration dataset, whereas all other loss functions cannot improve accuracy beyond the initialized quantization scheme.}
\label{fig:loss_types}
\end{figure}

We compare the infoNCE (contrastive) loss with other common  loss functions in \cref{fig:loss_types}. We find mean-squared error (MSE) to be equally (if not more) effective in the initial iterations of Evol-Q. However, as the number of passes grows, MSE does not perform as well as the infoNCE loss. Both cosine similarity and the Kullback–Leibler divergence (KL) fail to improve performance as the number of iterations increases. We postulate that the poor performance of these traditional loss functions is due to overfitting to the calibration dataset. On the other hand, the infoNCE loss is naturally regularized by the negative samples in the batch, allowing for it to preserve the quantization parameters that help discriminate between classes.

\subsection{Which layers contribute to non-smoothness?}
In \cref{fig:loss_land}, we visualize which layers of the self-attention mechanism yield the non-smooth loss curve. We show that learning the quantization scale for query, key, value (QKV) and projection layers is more difficult because their loss landscape is filled with local minima. We observe this property across different self-attention blocks and advocate for using ES to jump through the field of local minima.

\subsection{Layer-wise Weight Distributions}
\begin{figure}[h]
    \centering
    \includegraphics[width=1\linewidth] {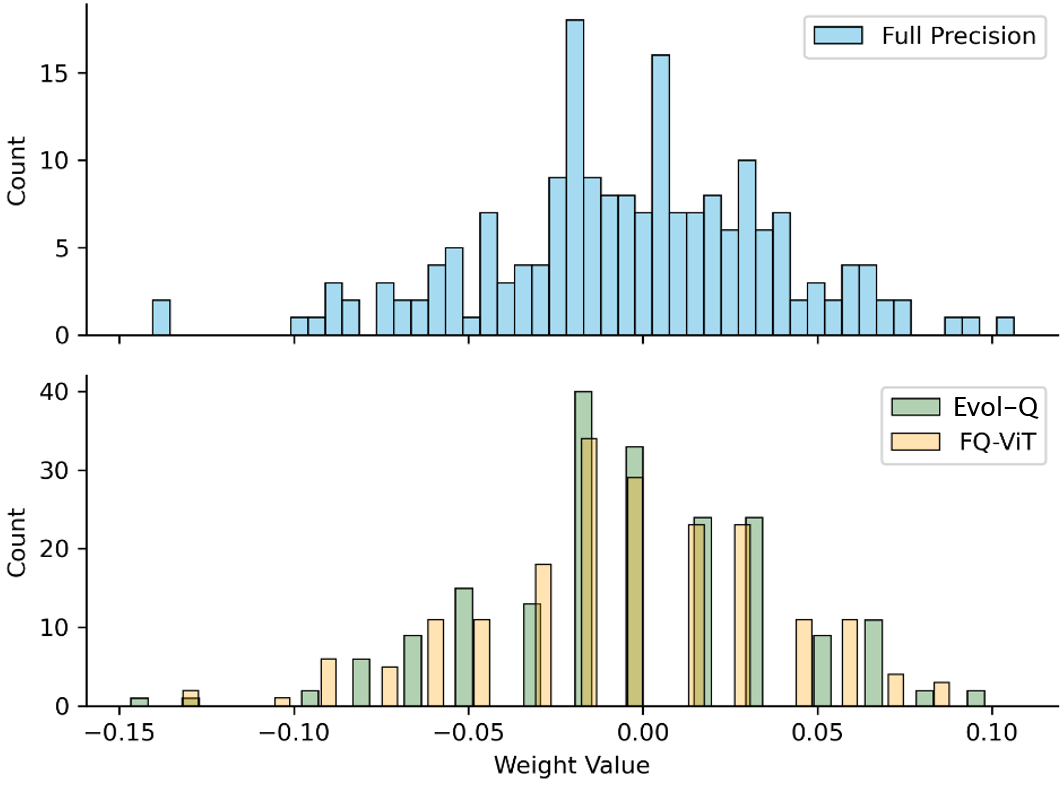}
    \caption{ The weight distribution of the projection layer for attention block \#1 of ViT-Base. The top plot shows the full precision weight distribution and the bottom plot shows the 8-bit weights using both FQ-ViT and Evol-Q.}
    \label{fig:mod1_hist}
\end{figure}

In \cref{fig:mod1_hist}, we compare the weight distributions of the full precision, FQ-ViT, and Evol-Q quantization schemes. Evol-Q's quantized values are only a small adjustment of FQ-ViT's, yet  Evol-Q has a $0.8\%$ improvement. In summary, a small adjustment in scale yields a significant boost in accuracy.

This observation is consistent with results in AdaRound~\cite{nagel2020up}, where authors show how choosing the correct rounding scheme can significantly impact performance. Unlike AdaRound~\cite{nagel2020up}, our method traverses the global loss landscape (rather than a layer-wise proxy) whereas AdaRound assumes a diagonal Hessian which does not hold in the ViT landscape.

We refer to Sec. F of the supplementary materials for more discussion on layer-wise distributions.

\subsection{Generalization to CNNs}
Our method begins with a pre-quantized model, $M_Q$, and adjusts the quantization scales to improve accuracy. We only require that the model can be abstracted into blocks, which makes our method readily applicable to other types of models such as CNNs, LSTMs, and Graph Neural Networks. In \cref{tab:cnns}, we show how Evol-Q is run on top of BRECQ to achieve state-of-the-art CNN quantization. In this case, our block is one convolutional layer and $\Delta$ is the stacked vector of quantization scales for the one layer's weight matrix. We find Evol-Q's method to be suitable for CNN quantization, achieving 1-2.5\% accuracy boost over BRECQ for 4-bit quantization.
We refer to supplementary material for an explanation of using the infoNCE loss to smooth out the CNN loss landscape.

\begin{table}
  \centering

  \begin{tabular}{|@{\hskip 0.05in}c@{\hskip 0.05in}|@{\hskip 0.05in}c@{\hskip 0.05in}|@{\hskip 0.05in}c@{\hskip 0.05in}|@{\hskip 0.05in}c@{\hskip 0.05in}|}
    \hline
     & ResNet-18 & ResNet-50 & RegNet-3.2GF\\
    \hline
    BRECQ [16] & 69.60 & 75.05  & 74.21  \\
    Evol-Q\textsuperscript{\textdagger} & \textbf{70.10} & \textbf{77.30} & \textbf{76.87}\\
    \hline
  \end{tabular}
  \footnotesize{\textsuperscript{\textdagger} Using BRECQ[16] as pre-quantized model $M_Q$}
  \caption{Top-1 Accuracy on ImageNet using 4W4A quantization.}
  \label{tab:cnns}
\end{table}
\subsection{Pareto Front for 8-bit Quantized ViTs}
\label{sec:qat}
\begin{figure}[h]
  \centering
    \includegraphics[width=0.85\linewidth]
           {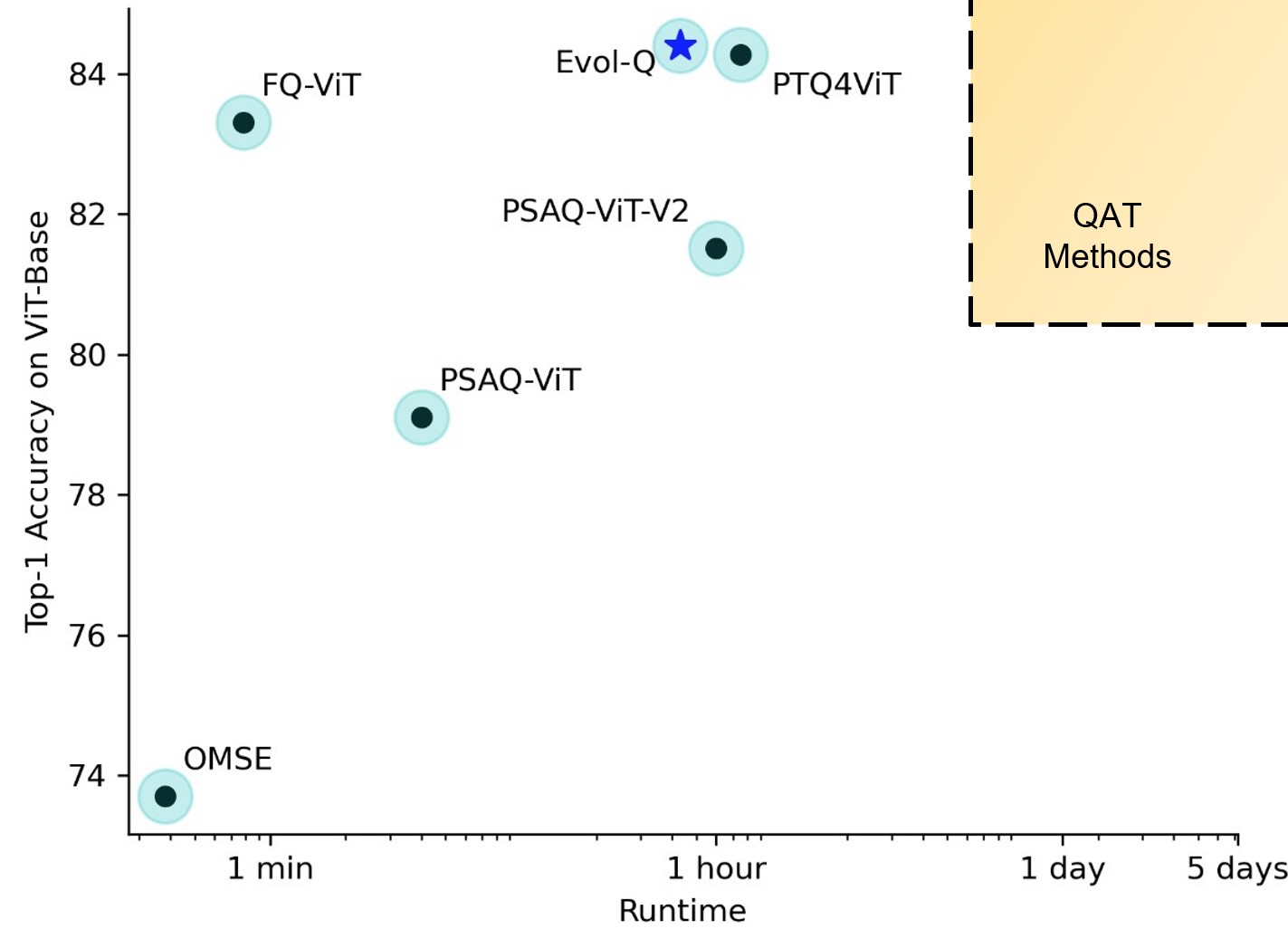}

  \caption{The runtime of 8-bit quantization methods for ViTs. We consider PTQ methods, and show the potential for QAT methods (orange square).}
  \label{fig:runtime}
\end{figure}

Evol-Q improves over existing PTQ techniques for vision transformers. In \cref{fig:runtime}, Evol-Q is on the Pareto front in terms of both top-1 accuracy and runtime for 8-bit ViT-Base. Current ViT-specific QAT methods~\cite{li2022q, xu2022tervit} do not report 8-bit accuracy, so we do not include them here. These QAT methods are likely to reach the Pareto front, but would take much longer than existing PTQ methods.

All open-source methods are run on a single Nvidia A100 GPU, but some code is not open-sourced at the time of submission. PSAQ-ViT-V2 does not report runtime, so we estimate it to be $60$ minutes based on PSAQ-ViT and the relative cost of additional steps.

\subsection{Runtime}
We run our method on an Nvidia A100-PCIE-40GB Tensor Core GPU and find that all experiments take less than one hour to run. The average runtime is shown in \cref{tab:runtime}. We use PyTorch 1.9.1, built with the CUDA 11.1 toolkit.

\begin{table}[h]
  \centering
  \begin{tabular}{c  c  c  c  c }
    \toprule
    & DeiT-T & DeiT-S & DeiT-B & ViT-B \\
    Runtime (m) & 41.5 & 46.3 & 41.6 & 43.2 \\

    \bottomrule
  \end{tabular}
  \caption{Evol-Q Runtime (in minutes) on one Nvidia A100 GPU}
  \label{tab:runtime}
\end{table}

In \cref{fig:runtime}, we compare runtime with other ViT quantization methods and demonstrate that our method achieves superior accuracy on ViT-Base. \cref{fig:runtime} only captures the Top-1 Accuracy of ViT-Base (or, alternatively, DeiT-base if ViT-Base is unavailable). We refer to the supplementary material for a wider discussion on how this plot changes with different models. 
\section{Conclusion}
Evol-Q achieves state-of-the-art results on ViT's highly non-smooth loss landscape with a high density of extremal points. Prior work on ViT quantization does not address the non-smooth loss landscape, nor how small perturbations in quantization scale can affect performance. Using evolutionary search and an infoNCE loss, Evol-Q evaluates small perturbations in quantization scale, improving accuracy by $\sim 0.5\%$ for 4-bit quantized vision transformers.

\section{Acknowledgements}

Partial work completed during a summer internship at Arm Ltd. A special thank you to Jesse Beu for overseeing this internship project, and to Feng Liang for helpful discussion. This work was sponsored by NSF CCF Grant No. 2107085 and the UT Cockrell School of Engineering Doctoral Fellowship.

\newpage

{\small
\bibliographystyle{ieee_fullname}
\bibliography{references}
}

\newpage
\input{supplementary}

\end{document}

%% file: authors.tex
\author{Natalia Frumkin\textsuperscript{\rm ~~1}~~~Dibakar Gope\textsuperscript{\rm ~~2}~~~Diana Marculescu\textsuperscript{\rm ~~1} \\
\textsuperscript{\rm 1} Chandra Family Department of Electrical and Computer Engineering,\\ The University of Texas at Austin \\
\textsuperscript{\rm 2} Arm, Inc.
}

%% file: supplementary.tex
\appendix
\maketitle
\ificcvfinal\thispagestyle{empty}\fi

\section{Using the infoNCE Loss}
The infoNCE loss is an effective self-supervised learning technique to learn intermediary representations, but why apply it to quantization?

\begin{figure}[h]
  \centering
  \begin{subfigure}{0.4\textwidth}
    \includegraphics[width=1\linewidth]
               {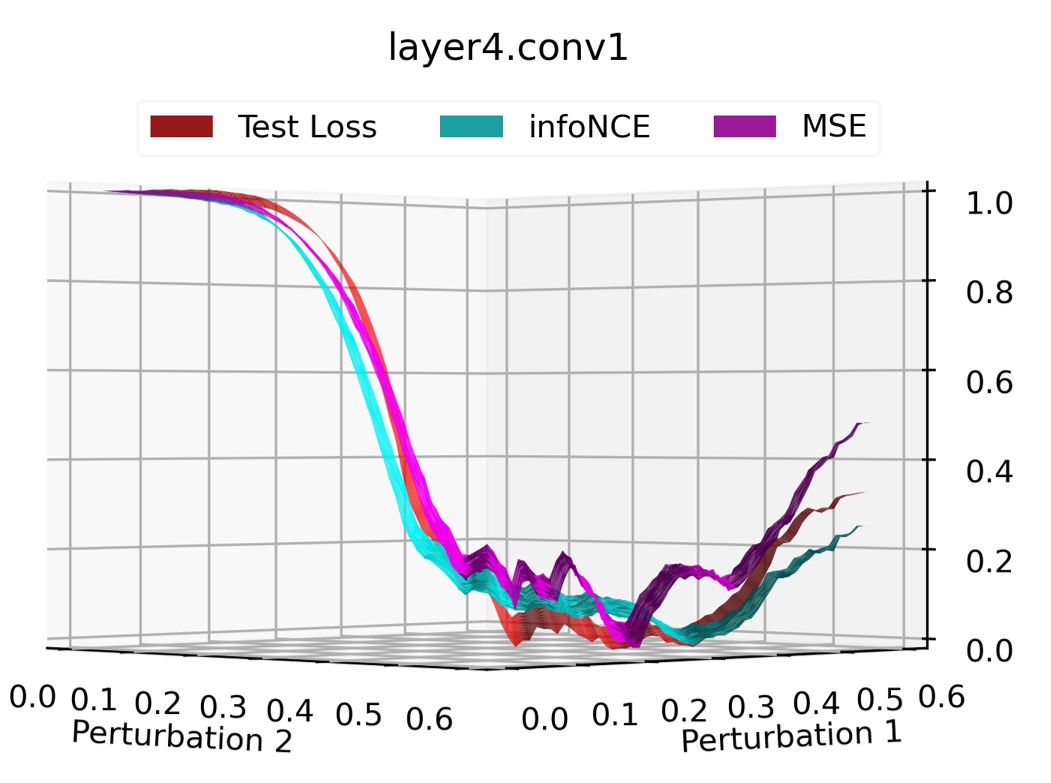}
    \caption{Comparison of the test loss landscape with MSE and infoNCE loss landscapes.}
    \label{fig:mse}
  \end{subfigure}
  \hfill
   \begin{subfigure}{0.4\textwidth}
    \includegraphics[width=1\linewidth]
               {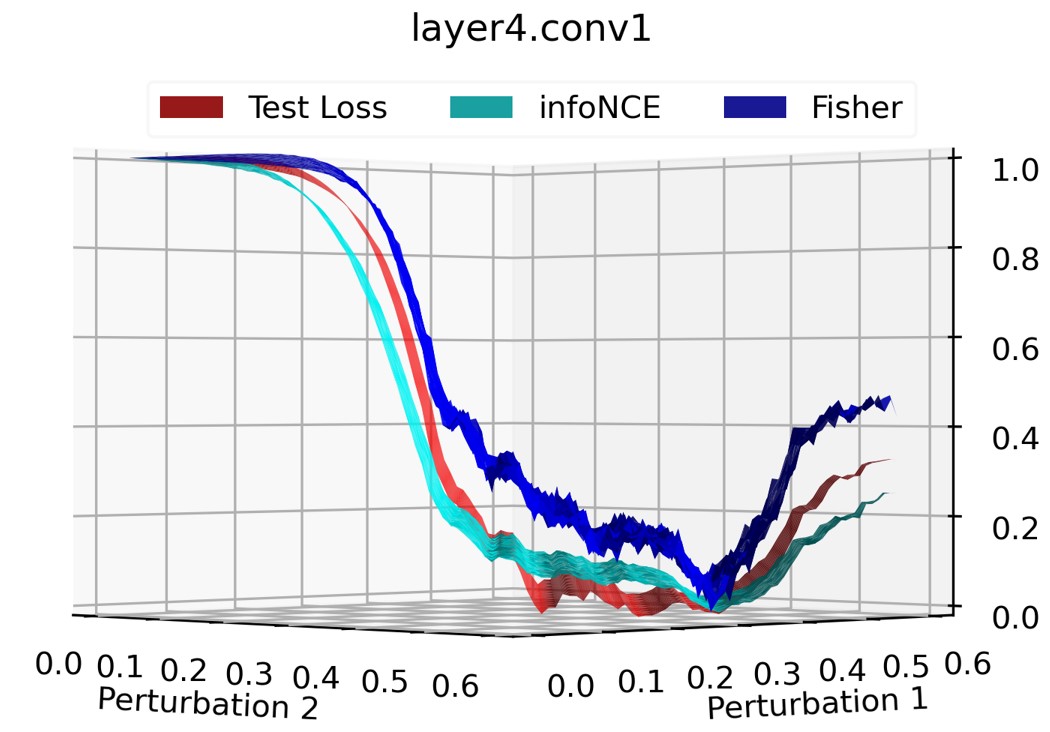}
    
    \caption{Comparison of the test loss landscape with Fisher and infoNCE loss landscapes.}
    \label{fig:fisher}
  \end{subfigure}
  \caption{Evaluating loss functions on ResNet-18. The infoNCE loss closely resembles the test loss (in red). In comparison, the MSE and Fisher loss are less smooth and do not accurately represent the test loss.}
  \label{fig:infoNCE-loss}
\end{figure}

We find, both experimentally (in Fig. 5 of main paper) and qualitatively in \cref{fig:infoNCE-loss}, that the infoNCE loss provides better results than existing loss functions for global quantization. To perform global quantization, we try to minimize a loss between the quantized and full precision outputs given by:

\begin{equation}
    \argmin_{\Delta}  
    \mathcal{L}\big( \hspace{1mm}x_{Q}(\Delta), \hspace{2mm} x_{FP} \hspace{1mm} \big)
\end{equation}

where $x_{Q}(\Delta)$ is the quantized prediction parameterized by quantization scales $\Delta$, and $x_{FP}$ is the full precision prediction. It may seem reasonable to use the mean-squared error (MSE) or cosine similarity as a loss function in this setup.
Unfortunately, PTQ methods only have access to a small calibration dataset, making it very easy for these loss functions to overfit to the few predictions available. The infoNCE loss combats this by using negative samples to encourage dissimilarity between  $x_Q$ and other predictions in the batch. We can see in \cref{fig:mse} that the infoNCE loss provides a smoothing effect when compared to the MSE loss. The infoNCE loss has a flatter minima which aids in generalization to the unknown test distribution.

Additionally, Hessian-based loss functions allow for second order gradient information, however, they must be estimated using some form of approximation such as the Fisher loss used in BRECQ~\cite{li2021brecq}. In \cref{fig:fisher}, we find the Fisher estimation to be noisy, and furthermore, does not accurately represent the underlying test loss landscape. The Fisher loss is an \textit{empirical estimation}, and is a poor approximation when the training distribution does not match the test data distribution~\cite{kunstner2019limitations}. We find that the infoNCE loss performs much better since it does not rely on any gradient approximation, and more closely resembles the test loss. In \cref{fig:fisher}, we can see that the infoNCE and Fisher losses share a similar minimum, but the infoNCE provides a flatter neighborhood around the minimum which is more robust to data distribution shift~\cite{keskar2016large,izmailov2018averaging}. As discussed above, the infoNCE loss encourages diversity of representations by encouraging dissimilarity between predictions.

\section{Ablation: Passes vs. Cycles}
\begin{figure}[h]
    \centering
    \includegraphics[width=0.9\linewidth]
           {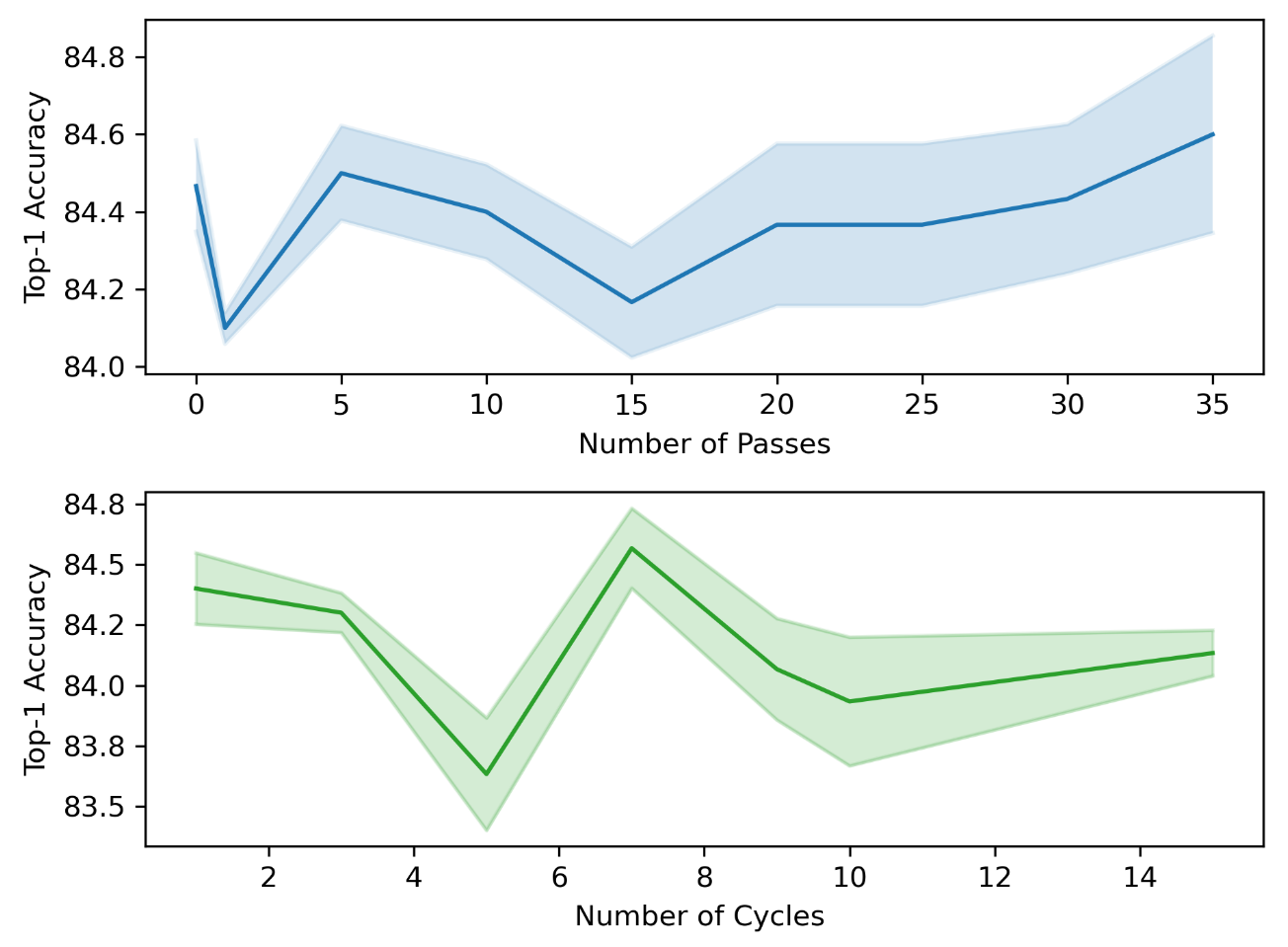}
    \caption{Ablation on number of cycles and passes.}
    \label{fig:pass_cycle_ablation}
\end{figure}

In \cref{fig:pass_cycle_ablation}, we ablate the number of passes, $P$, from $1$ to $35$. As we can see, a majority of the accuracy improvement occurs in the first $10$ passes, so we choose $P = 10$ for all experiments above. This allows for our method to run in less than one hour. However, we note that an additional accuracy boost may be enjoyed with more passes. 

We also ablate the number of cycles, $C$, to determine how many mutations should occur per block. We use $C = 3$ even though we see $C = 7$ is optimal in our ablation study. In practice, we find that the choice of $C$ is random seed and model dependent. We find that for some runs, the best choice is simply $1$ cycle, but in others it is $3$, $5$ or $7$. Ultimately, we choose $C = 3$ for consistency across experiments. 

\section{Ablation on Calibration Set Size}
\begin{figure}[h]
    \centering
    \includegraphics[width=0.9\linewidth]{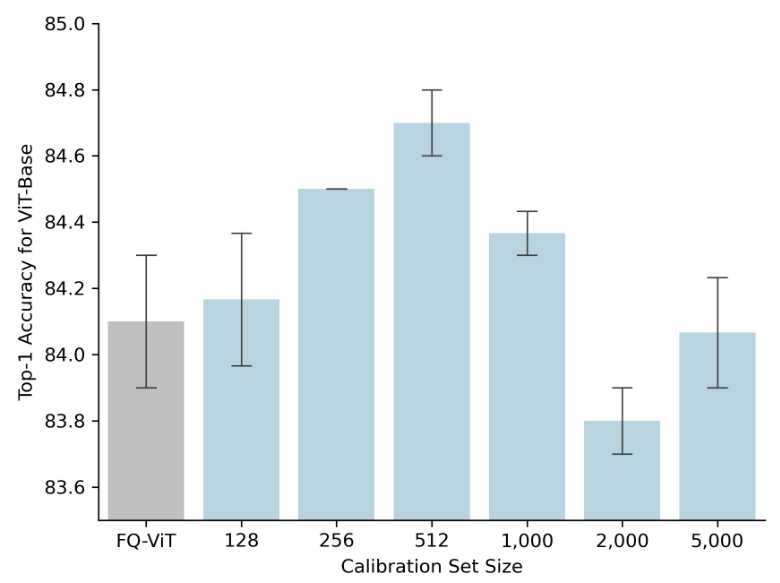}
    \caption{Ablation on Evol-Q's calibration dataset for sizes 128 to 5,000. }
    \label{fig:calib_size}
\end{figure}

As the calibration dataset increases, we'd expect better performance for our PTQ method. However, \cref{fig:calib_size} suggests that a 512 images yields the highest performance, whereas 2,000 and 5,000 images makes performance worse than FQ-ViT (which uses 1,000 calibration images). This is likely an artifact of the way we implement contrastive loss.

When we apply contrastive loss on a batch of images, the contrastive loss minimizes the distance to the corresponding full precision prediction, but maximizes the dissimilarity across all other images, regardless of the whether of not the other images are in the same class. Ideally, we want to avoid maximizing the dissimilarity within a class, so a smaller calibration dataset will minimize the likelihood of two images belonging to the same class.

We use 1,000 images in this paper as in prior work, however, accuracy may be improved by using only 512 images. Alternatively, a labelled calibration set may allow the contrastive loss to ignore other images belonging to the same class.

\section{On Variation across Random Seeds}
\begin{figure}[h]
    \centering
    \includegraphics[width=0.9\linewidth]
                   {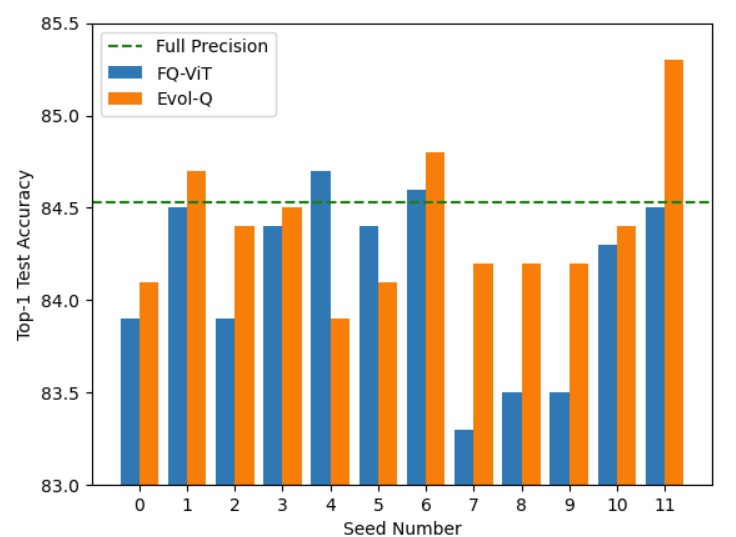}
    \caption{Comparing performance across 12 random seeds for 8W8A ViT-Base. 10/12 runs improve over the initial FQ-ViT quantization.}
    \label{fig:seeds}
    \vspace{-4mm}
\end{figure}

In \cref{fig:seeds}, we show the performance of Evol-Q compared to the baseline method, FQ-ViT. Across twelve random seeds, ten runs improve performance over FQ-ViT, and three result in top-1 accuracy that is superior to the full precision model. 

The random seed dictates which images are chosen for the calibration dataset, and we attribute the poor accuracy in seeds 4 and 5 to the poor choice of calibration set. This is a limitation of PTQ methods which rely on a calibration dataset, and so we employ a contrastive loss to combat overfitting (we can only minimize it's effect and not eliminate it).
\vspace{-3mm}

\begin{figure*}[h]
  \centering
    \includegraphics[width=1\linewidth]
           {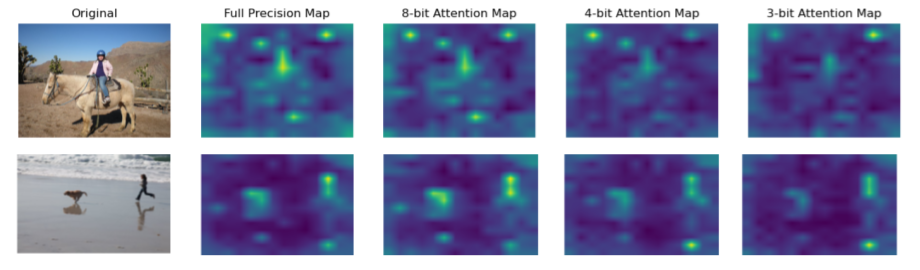}

  \caption{Attention maps for different quantization levels. Evol-Q's quantized models preserve the spacial locality of the full precision feature map. As the quantization level becomes more extreme, the attention map becomes subject to decreased resolution.}
  \label{fig:att_maps}
\end{figure*}

\section{Impact on Attention Maps}
We find that Evol-Q preserves the spatial integrity of the full precision feature maps even as quantization forces discretization of the attention mechanism. In \cref{fig:att_maps}, as quantization becomes more severe from 8-bit to 3-bit, the resolution of the feature map degrades, as is expected when only a finite number of values can be expressed in the quantized scheme. This attention map visualization is averaged over all blocks, and serves as qualitative inspection of how the quantized network's attention mechanism is performing. All in all, \cref{fig:att_maps} provides confidence that Evol-Q's quantized attention maps learn reasonable representations of the original full precision network.

\section{Layer-wise Weight Distributions}
The weight distributions for ViT-Base's projection layers are shown in \cref{fig:weight_dists}. To recap, the projection layer is the final linear layer of each attention block\footnote{$W^O$ in Pytorch's \texttt{torch.nn.MultiheadAttention()}}. 

The beauty of Evol-Q is in its global optimization strategy -- learning quantization scales with respect to a global objective allows Evol-Q to choose scales for the intermediary layers which improve quantization for other layers. FQ-ViT may approximate the full precision weight distribution well, however, a matching layer-wise distribution may not translate to overall performance gain. As explained in the main paper, a small perturbation in quantization scale can reap a huge accuracy gain. We can see that Evol-Q's layer-wise distributions are not very different than FQ-ViT, yet Evol-Q has a 0.15\% accuracy improvement over FQ-ViT for ViT-Base. In summary, we find that Evol-Q's slight adjustment in quantization scale can greatly improve accuracy.

Please refer to the last page for \cref{fig:weight_dists}.

\begin{figure}[h]
    \centering
    \includegraphics[width=0.9\linewidth]{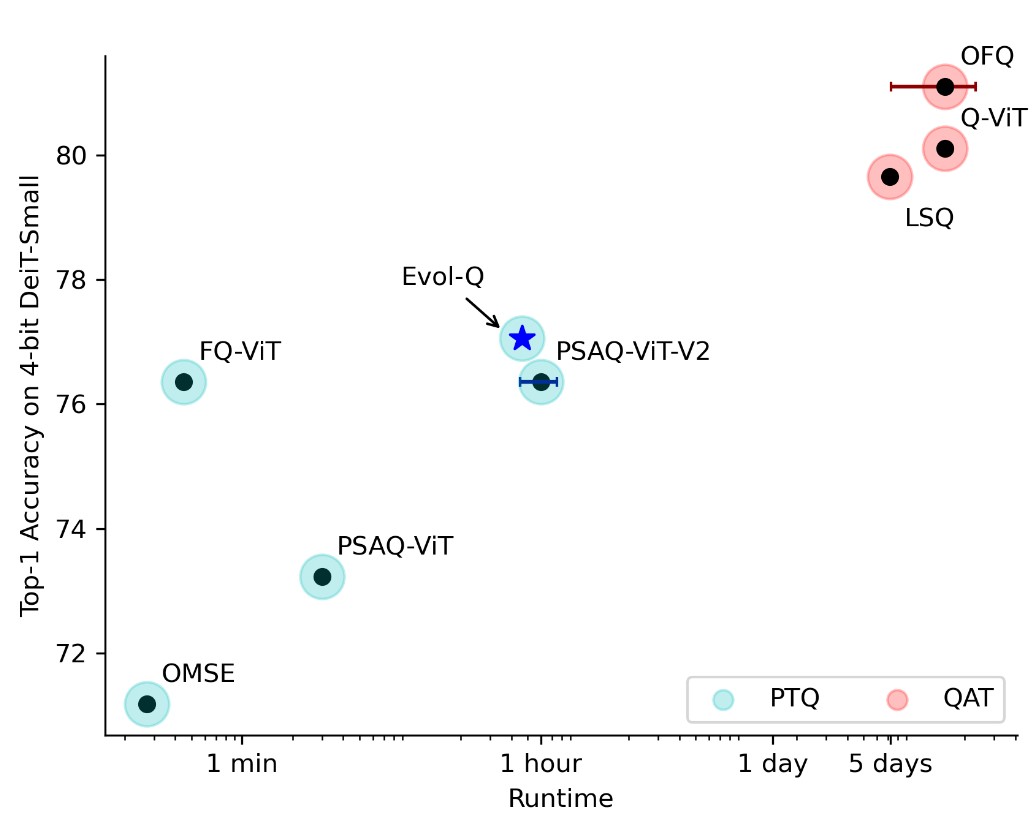}
    \caption{Runtime \textit{vs.} Accuracy for 4-bit DeiT-Small using existing vision transformer techniques. We compare PTQ methods (blue) and QAT methods (red) on the same plot and show that Evol-Q is on the Pareto front. We estimate runtime for PSAQ-ViT-V2~\cite{li2022psaq} and OFQ~\cite{liu2023oscillation}, and indicate uncertainty using error bars.}
    \label{fig:runtime_v_acc}
\end{figure}

\begin{figure*}[b]
  \centering
  \begin{subfigure}{0.3\linewidth}
    \includegraphics[width=1\linewidth]
               {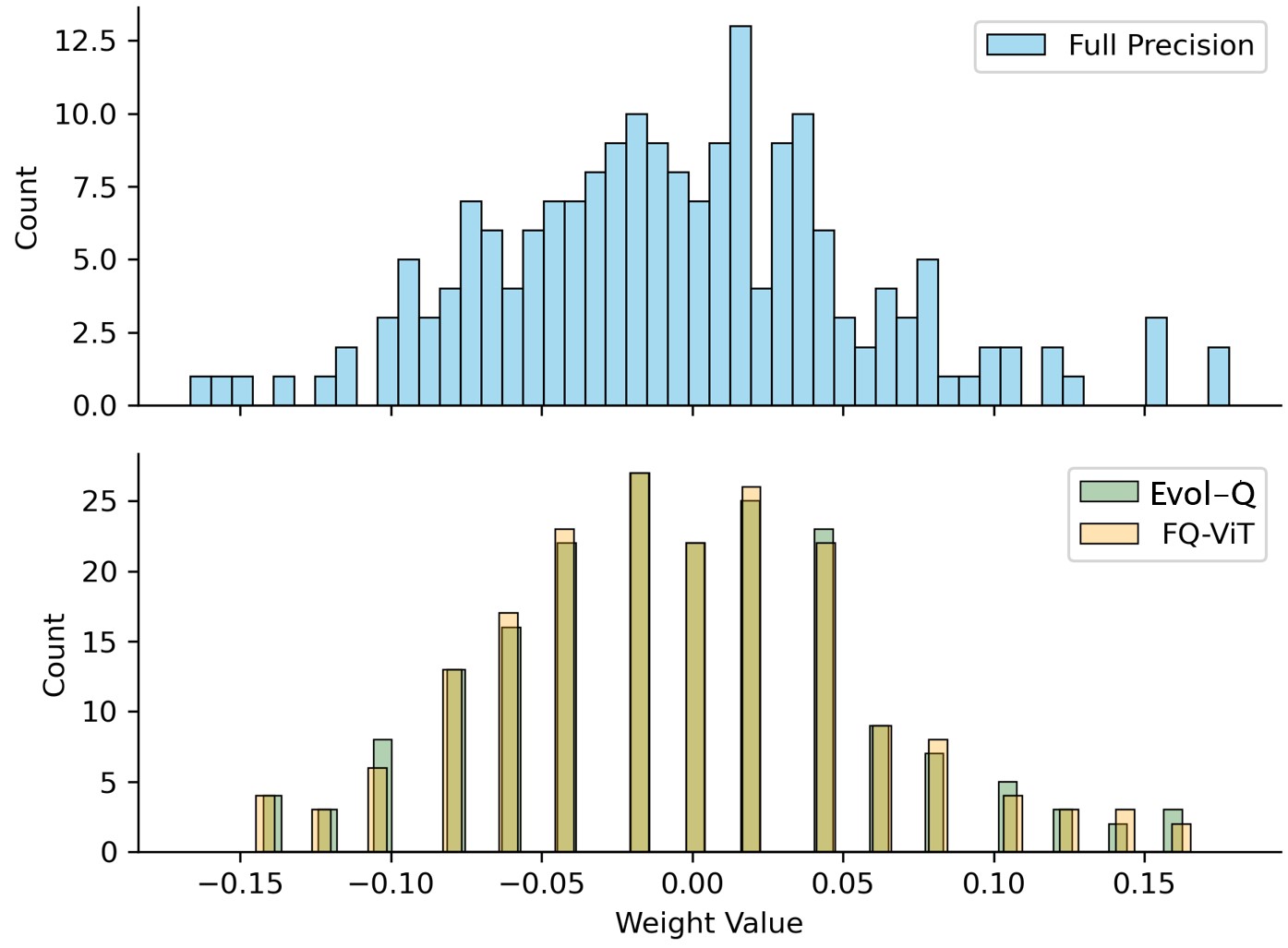}
    \caption{Block \#0}
    \label{fig:mod0}
    \vspace{2mm}
  \end{subfigure}
  \hfill
  \begin{subfigure}{0.3\linewidth}
    \includegraphics[width=1\linewidth]
               {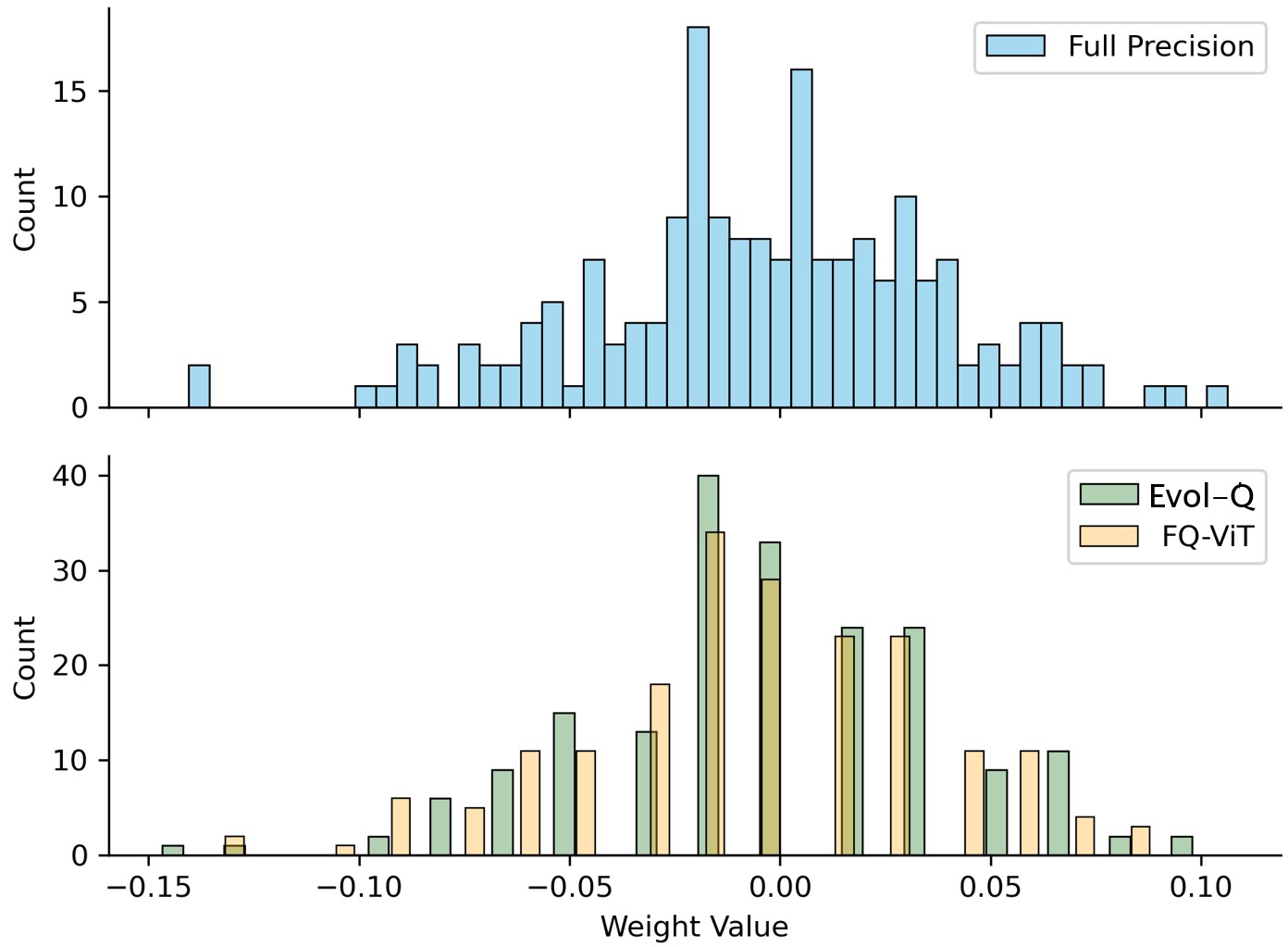}
    \caption{Block \#1}
    \label{fig:mod1}
    \vspace{2mm}
  \end{subfigure}
  \hfill
    \begin{subfigure}{0.3\linewidth}
    \includegraphics[width=1\linewidth]
               {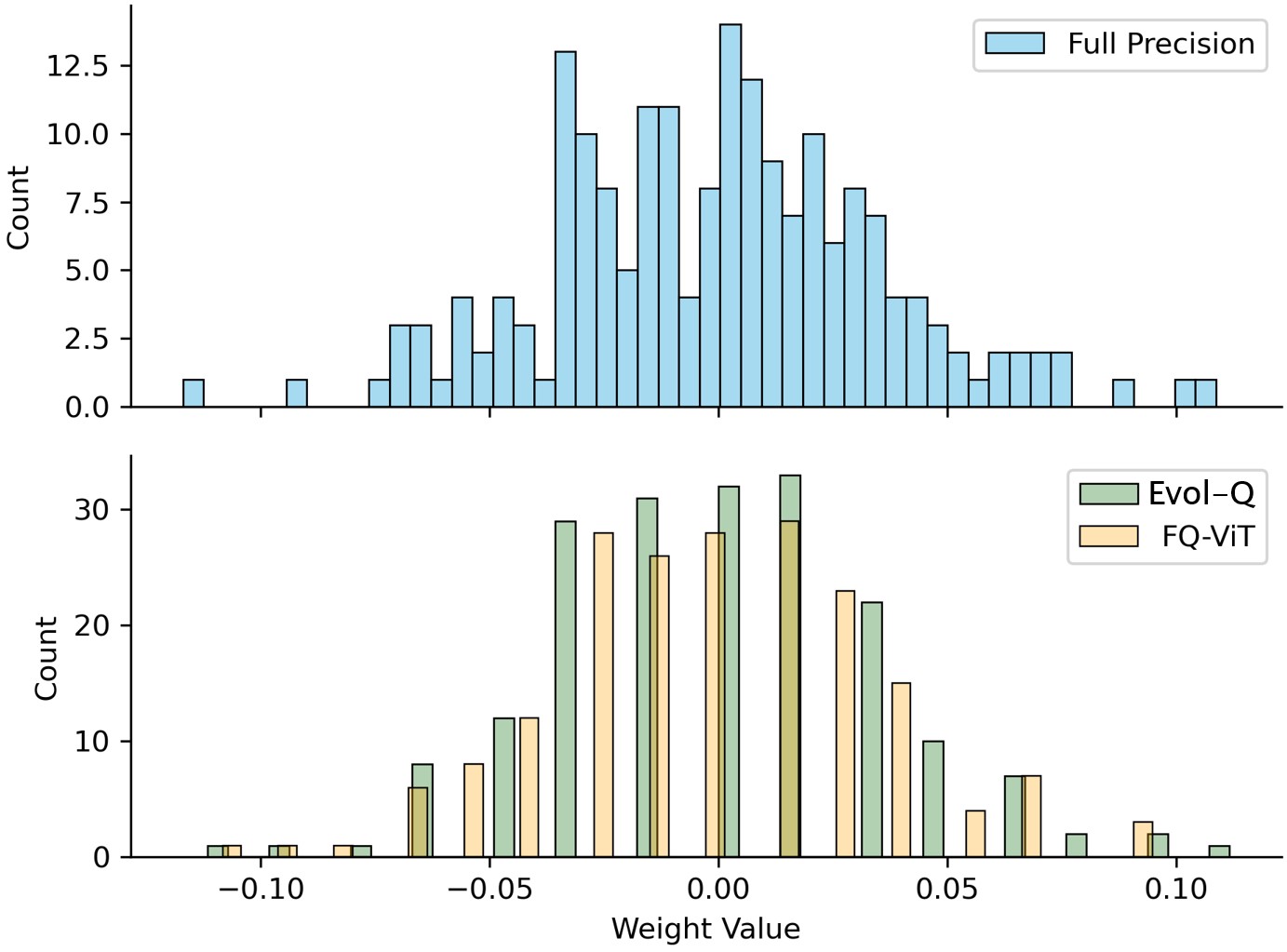}
    \caption{Block \#2}
    \label{fig:mod2}
    \vspace{2mm}
  \end{subfigure}
  \begin{subfigure}{0.3\linewidth}
    \includegraphics[width=1\linewidth]
               {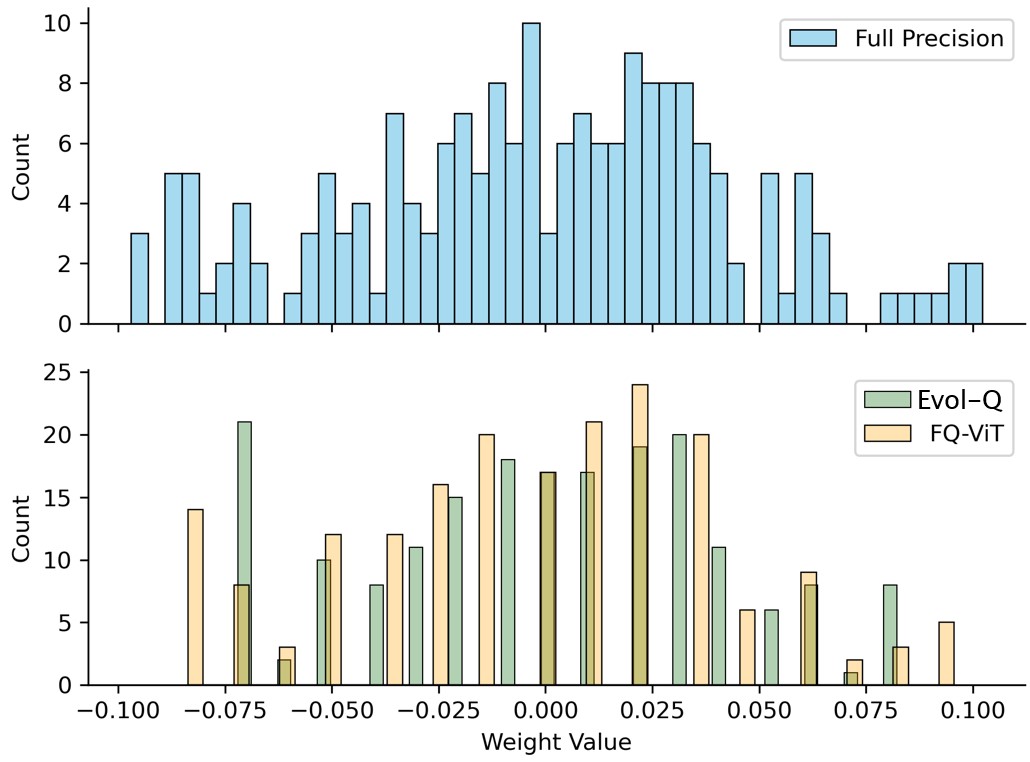}
    \caption{Block \#3}
    \label{fig:mod3}
    \vspace{2mm}
  \end{subfigure}
  \hfill
  \begin{subfigure}{0.3\linewidth}
    \includegraphics[width=1\linewidth]
               {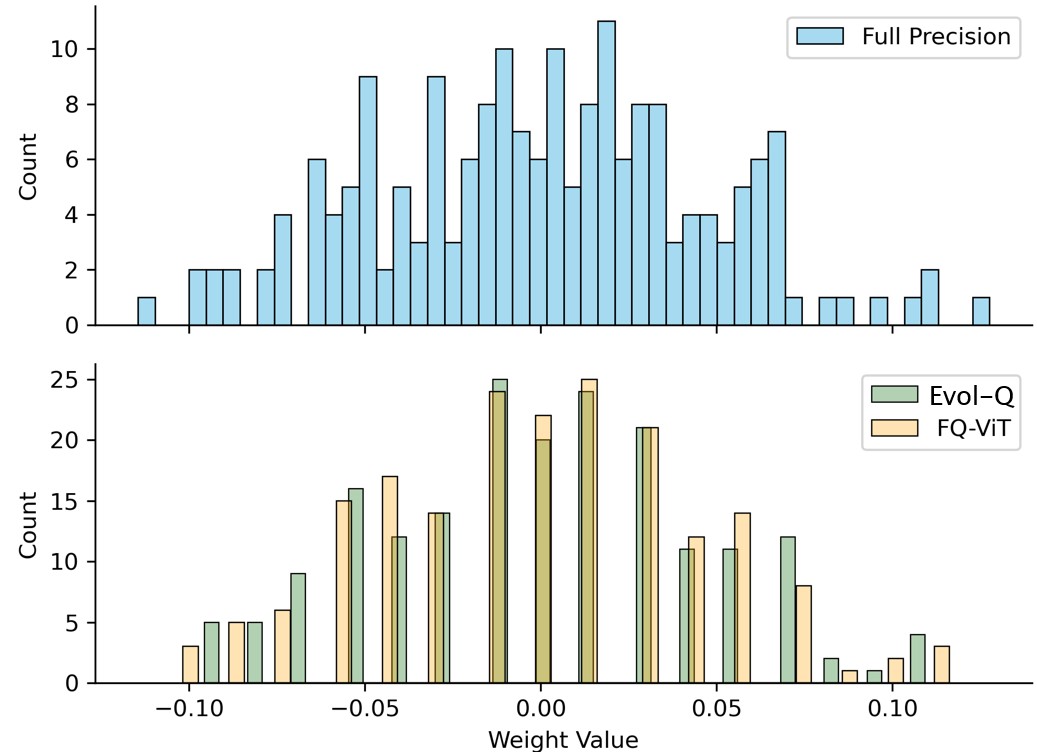}
    \caption{Block \#4}
    \label{fig:mod4}
    \vspace{2mm}
  \end{subfigure}
  \hfill
    \begin{subfigure}{0.3\linewidth}
    \includegraphics[width=1\linewidth]
               {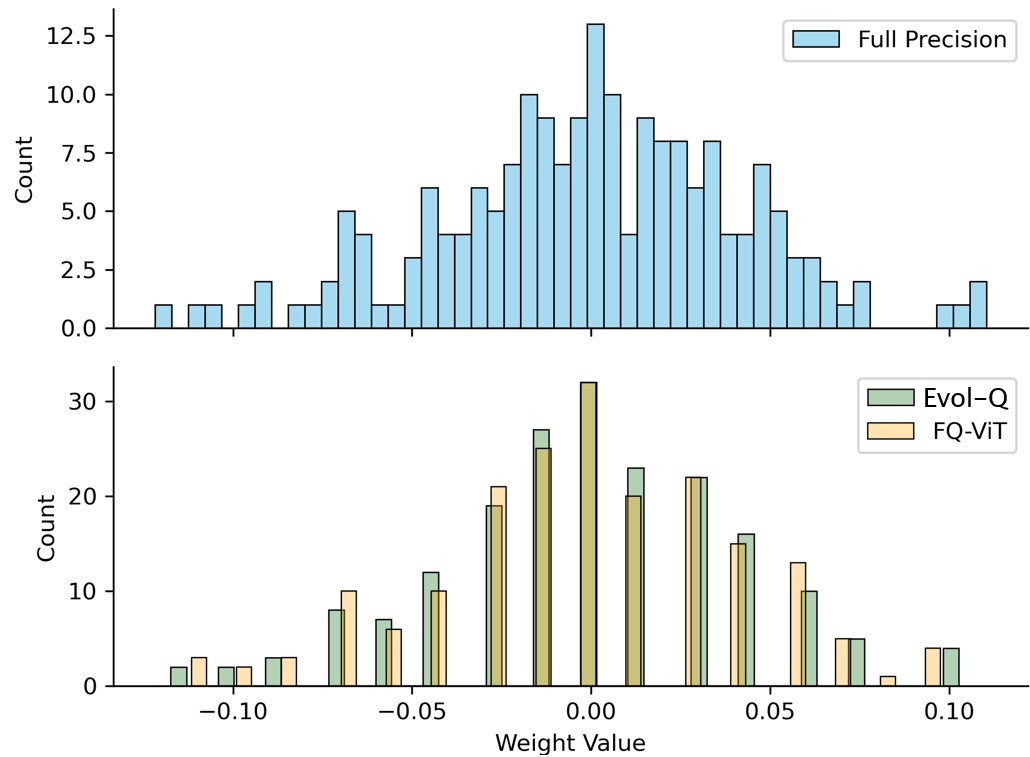}
    \caption{Block \#5}
    \label{fig:mod5}
    \vspace{2mm}
  \end{subfigure}
    \vspace{2mm}
    \begin{subfigure}{0.3\linewidth}
    \includegraphics[width=1\linewidth]
               {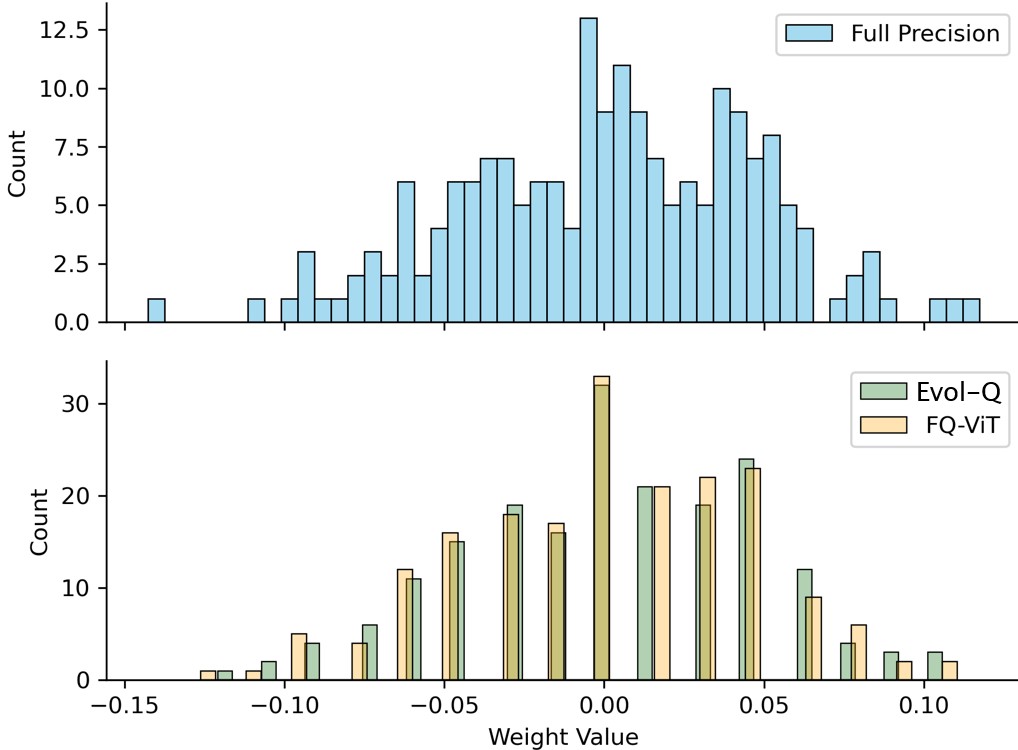}
    \caption{Block \#6}
    \label{fig:mod6}
  \end{subfigure}
  \hfill
  \begin{subfigure}{0.3\linewidth}
    \includegraphics[width=1\linewidth]
               {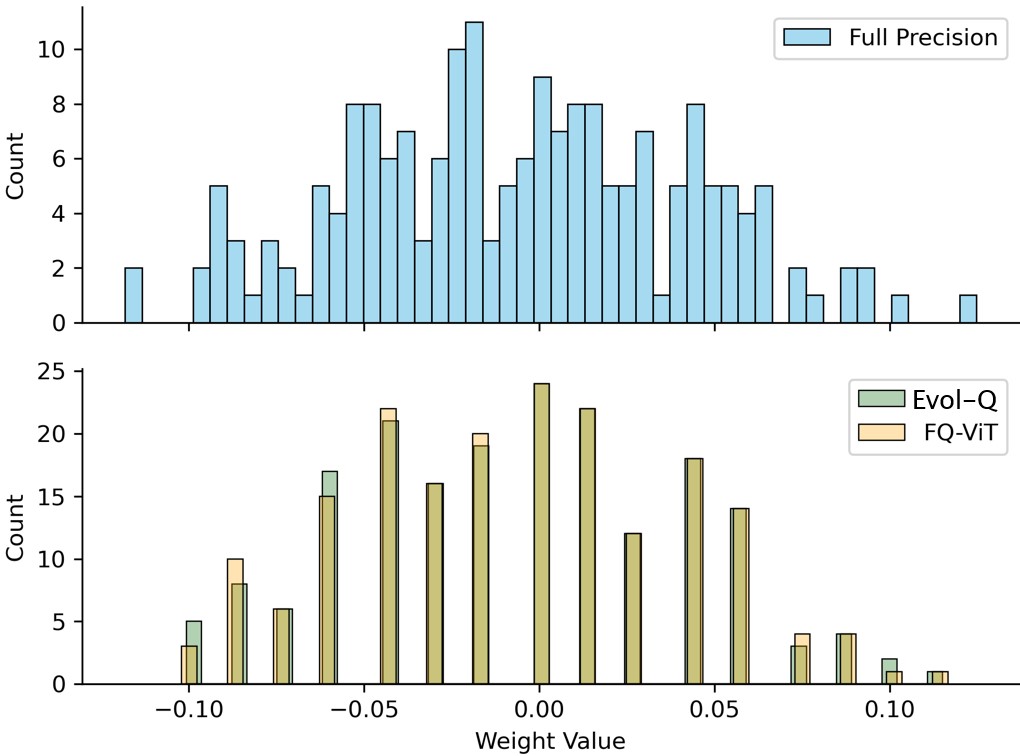}
    \caption{Block \#7}
    \label{fig:mod7}
  \end{subfigure}
  \hfill
    \begin{subfigure}{0.3\linewidth}
    \includegraphics[width=1\linewidth]
               {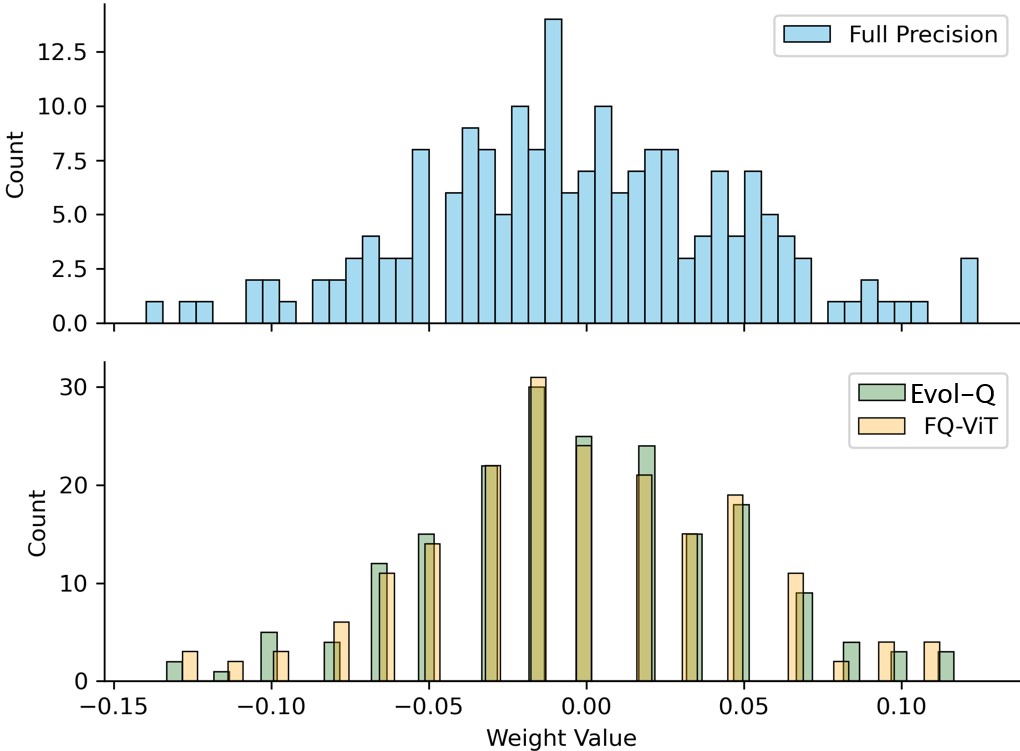}
    \caption{Block \#8}
    \label{fig:mod8}
  \end{subfigure}

    \begin{subfigure}{0.3\linewidth}
    \includegraphics[width=1\linewidth]
               {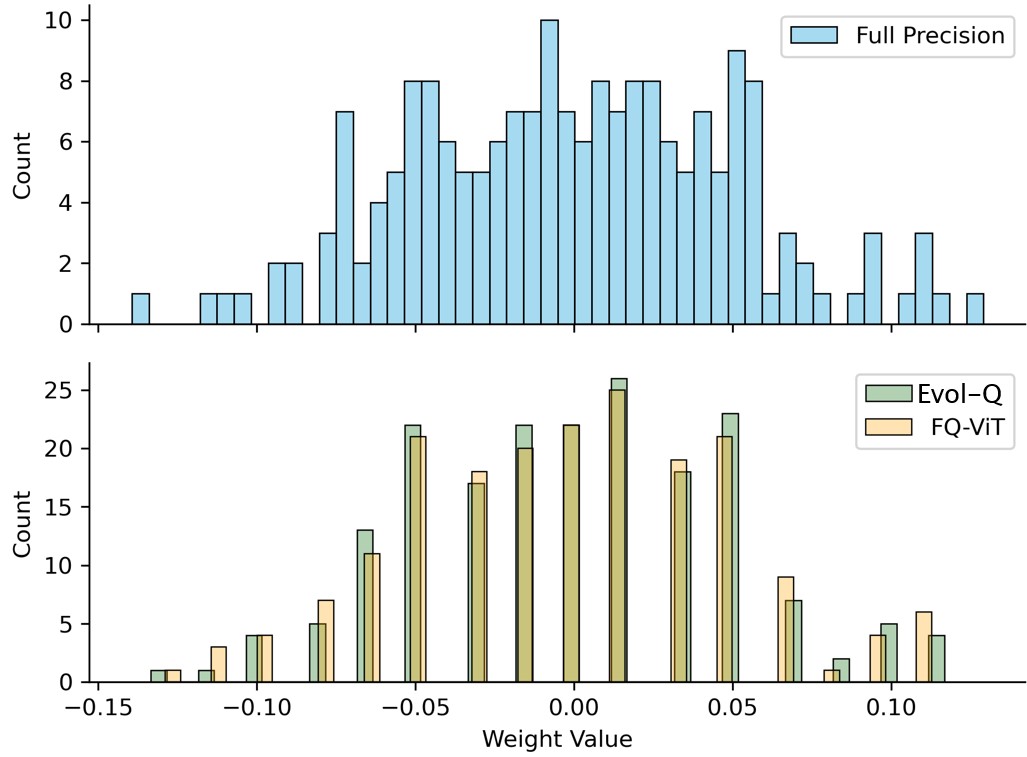}
    \caption{Block \#9}
    \label{fig:mod9}
  \end{subfigure}
  \hfill
  \begin{subfigure}{0.3\linewidth}
    \includegraphics[width=1\linewidth]
               {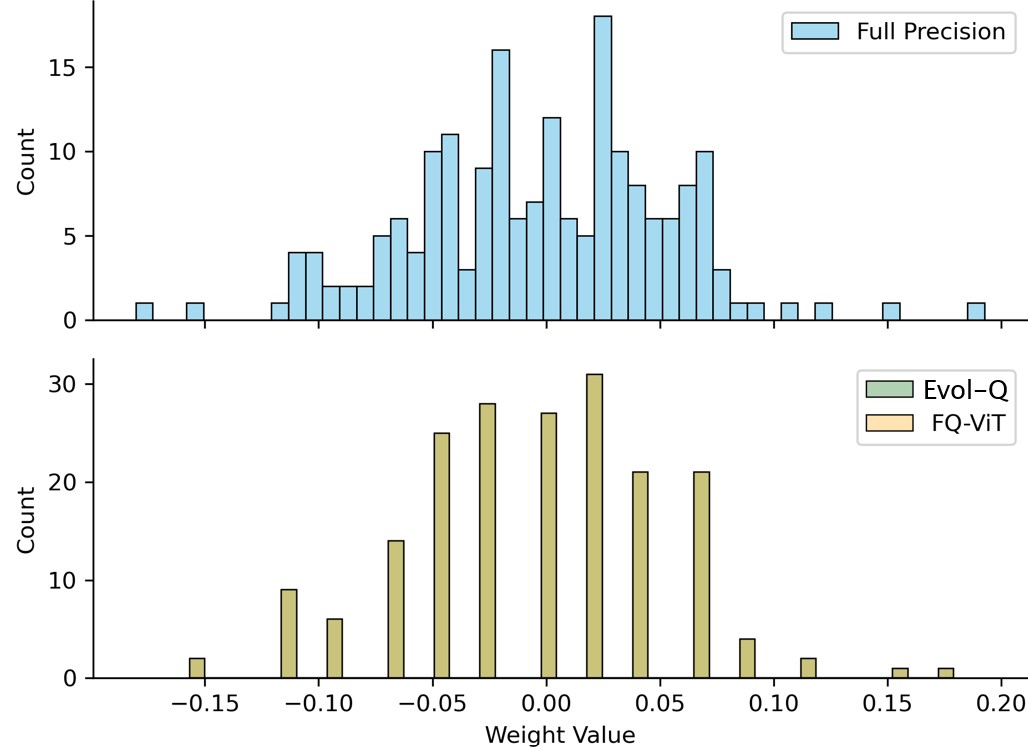}
    \caption{Block \#10}
    \label{fig:mod10}
  \end{subfigure}
  \hfill
    \begin{subfigure}{0.3\linewidth}
    \includegraphics[width=\linewidth]
               {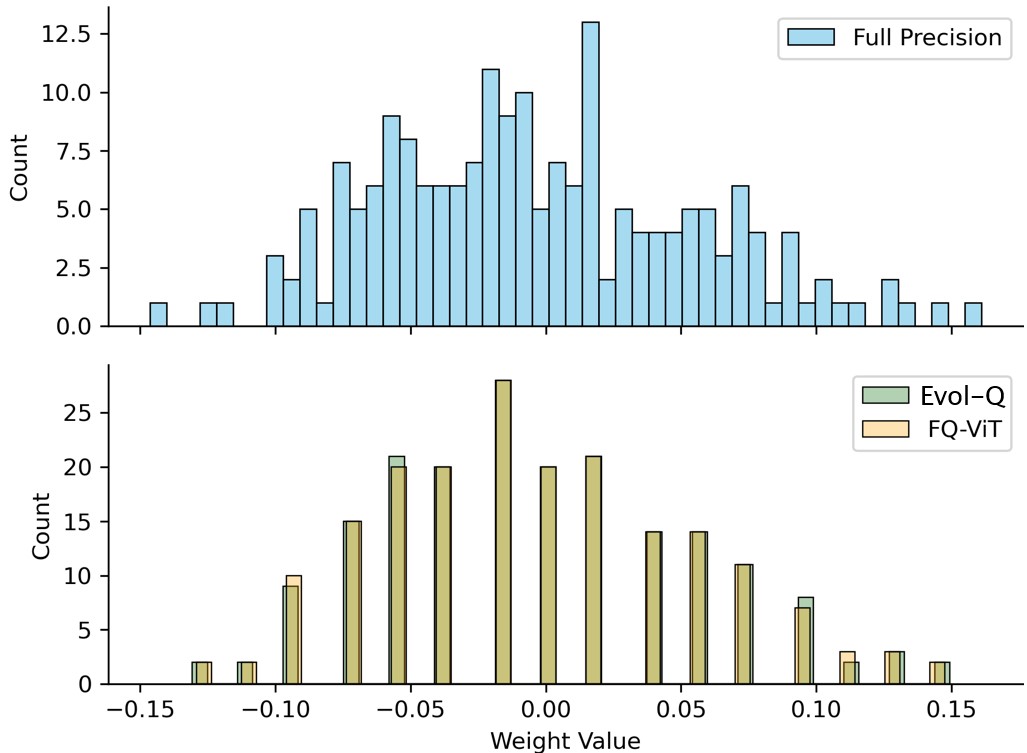}
    \caption{Block \#11}
    \label{fig:mod11}
  \end{subfigure}

  \caption{Weight distributions for the projection layers of all attention blocks for ViT-Base. The 12 blocks are numbered from 0-11, with block \#1 being the same as reported in Fig. 4 of the main paper. Evol-Q (green) has a 0.15\% Top-1 accuracy improvement over FQ-ViT (yellow). }
  \label{fig:weight_dists}
\end{figure*} 

\begin{table}
\begin{subtable}{\linewidth}
  \centering
  \rowcolors{2}{gray!25}{white}
  \begin{tabular}{| c |  c  c  c  c |}
    \hline
    \multicolumn{5}{|c|}{3-bit weights, 8-bit activations (3W8A)} \\
    \hline
    Method &  DeiT-T & DeiT-S & DeiT-B & ViT-B \\
    \hline
    \multicolumn{1}{|l|}{FQ-ViT} &  35.79 & 60.58 & 72.11 & 55.33 \\
    + OMSE & 52.03 & 65.27 & 75.00 & 62.83 \\
    \hspace{2mm}+ Bias Corr & 56.17 & 68.53 & 77.57 & 73.27 \\
    Evol-Q (ours)  & \textbf{58.93} & \textbf{69.93} & \textbf{78.40} & \textbf{75.00} \\

    \hline
  \end{tabular}
    \caption{3-bit weights, 8-bit activations}

  \label{tab:3w8a_supp}
\end{subtable}

\begin{subtable}{\linewidth}
  \centering
  \rowcolors{2}{gray!25}{white}
  \begin{tabular}{| c |  c  c  c  c |}
    \hline
    \multicolumn{5}{|c|}{4-bit weights, 8-bit activations (4W8A)} \\
    \hline
    Method &  DeiT-T & DeiT-S & DeiT-B & ViT-B \\
    \hline
    \multicolumn{1}{|l|}{FQ-ViT} &  66.91 & 76.93 & 79.99 & 78.73 \\
    + OMSE & 66.03 & 77.17 & 80.30 & 78.90 \\
    \hspace{2mm}+ Bias Corr & 67.27 & 78.03 & 80.43 & 79.37 \\
    Evol-Q (ours)  & \textbf{68.47} & \textbf{78.30} & \textbf{81.07} & \textbf{80.37} \\
    \hline
  \end{tabular}
    \caption{4-bit weights, 8-bit activations}
  \label{tab:4w8a_supp}
\end{subtable}

\begin{subtable}{\linewidth}
  \centering
  \rowcolors{2}{gray!25}{white}
  \begin{tabular}{| c |  c  c  c  c |}
    \hline
    \multicolumn{5}{|c|}{8-bit weights, 8-bit activations (8W8A)} \\
    \hline
    Method &  DeiT-T & DeiT-S & DeiT-B & ViT-B \\
    \hline
    \multicolumn{1}{|l|}{FQ-ViT} &  71.61 & 79.17 & 81.20 & 83.30 \\
    + OMSE & 72.17 & 80.30 & 82.17 & 82.47 \\
    \hspace{2mm}+ Bias Corr & 72.33 & 79.87 & 82.07 & 82.43 \\
    Evol-Q (ours)  & \textbf{72.37} & \textbf{80.33} & \textbf{82.47} & \textbf{84.40} \\
    \hline
  \end{tabular}
  \caption{8-bit weights, 8-bit activations}
  \label{tab:8w8a_supp}
\end{subtable}
  \caption{We add OMSE quantization and Bias Correction (Bias Corr) on top of FQ-ViT. Finally, we apply Evol-Q on top of all three methods to achieve state-of-the-art PTQ quantization. We show results for 3W8A, 4W8A, 8W8A in \cref{tab:3w8a_supp}, \cref{tab:4w8a_supp}, and \cref{tab:8w8a_supp} respectively.}
  \label{tab:add_omse_bc}
\end{table}

\section{Pareto Front for 4-bit DeiT-Small}
Since most methods report 4-bit weights for DeiT-Small, we compare these methods in terms of both runtime \& accuracy. In \cref{fig:runtime_v_acc} we illustrate tradeoff between runtime and accuracy for PTQ and QAT methods. In comparison to 8-bit ViT-Base (Fig. 7 in the main paper), this figure includes QAT results which are unavailable in the 8-bit setting. We estimate runtime for PSAQ-ViT-V2~\cite{li2022psaq} and OFQ~\cite{liu2023oscillation}, since they do not open-source their code, nor report runtime. Evol-Q is on the Pareto curve (note x-axis is log scale), and has the best accuracy of all PTQ methods. Still, there is a performance gap ($\sim 2.5-3\%$) when compared to QAT methods, illustrating that there is room to improve for PTQ methods.

\section{Adding Bias Correction and OMSE}
OMSE quantization~\cite{choukroun2019low} and Bias Correction~\cite{banner2019post} are statistical techniques we can use to improve quantization performance. We apply them on the original FQ-ViT model, and then use Evol-Q to achieve state-of-the-art PTQ performance. In \cref{tab:add_omse_bc} (last page), we can see the benefits of applying OMSE and Bias Correction techniques and how adding Evol-Q on top of these can boost performance even more. 

In this paper, we have shown how Evol-Q can boost performance in a variety of scenarios and does not require a cherry-picked setting. We show that Evol-Q works on top of BRECQ for CNNs, FQ-ViT for ViTs, and even works in this setting, where we boost FQ-ViT's accuracy by adding Bias Correction and OMSE.

In summary, we are confident that Evol-Q's novel optimization method in conjunction with evaluating small scale perturbations  is orthogonal to other quantization methods and can be used in a variety of scenarios to improve accuracy.